\documentclass{article}

\usepackage[preprint]{neurips_2024}

\usepackage[utf8]{inputenc} 
\usepackage[T1]{fontenc}    
\usepackage{hyperref}       
\usepackage{url}            
\usepackage{booktabs}       
\usepackage{amsfonts}       
\usepackage{nicefrac}       
\usepackage{microtype}      
\usepackage{xcolor}         

\usepackage{graphicx}
\usepackage{subfigure}
\usepackage{tikz}
\usepackage{csquotes}
\usepackage{hyperref}

\usepackage{amsmath}
\usepackage{amssymb}
\usepackage{mathtools}
\usepackage{amsthm}
\usepackage{algorithm}
\usepackage{algorithmic}
\usepackage{diagbox}
\usepackage{wrapfig}
\usepackage{tikz}
\usepackage{tcolorbox}

\title{In-Context Exploiter for Extensive-Form Games}

\author{%
  Shuxin Li\thanks{Equal contribution} \\
  Nanyang Technological University\\
  \texttt{shuxin.li@ntu.edu.sg} \\
  \And 
  Chang Yang\footnotemark[1] \\
  The Hong Kong Polytechnic University\\
  \texttt{chang.yang@connect.polyu.edu.hk} \\
  \And 
  Youzhi Zhang \\
  CAIR, HKISI, CAS \\
  \texttt{youzhi.zhang@cair-cas.org.hk} \\
  \And 
  Pengdeng Li \\
  Nanyang Technological University\\
  \texttt{pengdeng.li@ntu.edu.sg} \\
  \And 
  Xinrun Wang\thanks{Corresponding Author}\\
  Singapore Management University\\
  \texttt{xrwang@smu.edu.sg} \\
  \And 
  Xiao Huang \\
  The Hong Kong Polytechnic University\\
  \texttt{xiaohuang@comp.polyu.edu.hk} \\
  \And 
  Hau Chan \\
  University of Nebraska\\
  \texttt{hchan3@unl.edu} \\
  \And 
  Bo An \\
  Nanyang Technological University\\
  \texttt{boan@ntu.edu.sg} \\
}

\begin{document}

\maketitle

\begin{abstract}
Nash equilibrium (NE) is a widely adopted solution concept in game theory due to its stability property. However, we observe that the NE strategy might not always yield the best results, especially against opponents who do not adhere to NE strategies. Based on this observation, we pose a new game-solving question: \emph{Can we learn \textbf{a model} that can exploit \textbf{any, even NE, opponent} to maximize their own utility?}
In this work, we make the first attempt to investigate this problem through in-context learning. Specifically, we introduce a novel method, In-Context Exploiter (ICE), to train a single model that can act as any player in the game and adaptively exploit opponents entirely by in-context learning. Our ICE algorithm involves generating diverse opponent strategies, collecting interactive history training data by a reinforcement learning algorithm, and training a transformer-based agent within a well-designed curriculum learning framework. Finally, comprehensive experimental results validate the effectiveness of our ICE algorithm, showcasing its in-context learning ability to exploit any unknown opponent, thereby positively answering our initial game-solving question. 
\end{abstract}

\section{Introduction}
The domain of game-solving has consistently served as a benchmark for the advancement in Artificial Intelligence (AI) \citep{tammelin2015solving,moravvcik2017deepstack}. It tests the boundaries of the strategic reasoning and decision-making capabilities of AI systems. It is well known that in game theory, Nash equilibrium (NE) \citep{nash1950equilibrium} is the standard solution concept, which describes a situation where no player can increase their utility by unilaterally deviating. In many security-related cases, the NE strategy plays a critical role since it is the most conservative strategy that can achieve the best performance in the worst-case scenario \citep{jain2011double,jain2013security}.
However, the NE strategy may not achieve the best utility in some scenarios where the opponents do not play the NE strategy. For example, in the classic game of rock-paper-scissors, an opponent adhering strictly to an NE strategy would randomize their selections, ensuring no discernible pattern emerges. If an adversary were to deviate from this randomness and favor one choice, a non-NE strategy that exploits this bias could yield a better outcome (a detailed explanation can be found in Sec.~\ref{example}). 

\begin{figure}[t]
    \centering
\includegraphics[width=0.6\columnwidth]{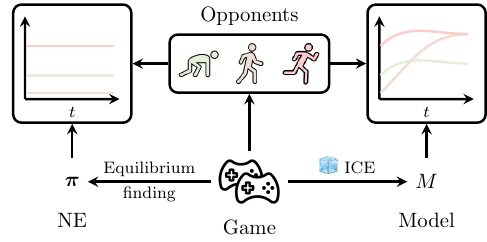}
    \caption{Comparison between equilibrium finding and our method}
    \label{fig:1}
\end{figure}
This brings us to the core question of game-solving: 
\begin{displayquote}
\emph{Can we learn \textbf{a model} which can exploit \textbf{any opponent} to maximize his own utility?}    
\end{displayquote}
Put simply, is it possible to consistently obtain the best utility by exploiting any type of opponent, even when we have no prior knowledge of their strategy? 

The field of opponent exploitation has seen significant advancements. \citet{foerster2017learning} introduced a method for deducing the parameters of opponents’ policies based on historical interaction data. However, this approach requires extensive training data to accurately adapt to new opponents.
Recently, \citet{wu2022l2e} developed a deep learning framework for implicit opponent modeling, called Learning-to-Exploit (L2E) which includes an adversarial training procedure to autonomously generate opponents, thereby reducing the training data requirements.
However, these methods are not suited to address the game-solving problem we propose, since they necessitate retraining upon encountering a new opponent. This limitation points to a lack of generalizability, meaning they cannot exploit any opponent only using a model without parameter updating.

To solve these issues, we resort to in-context learning which has gained significant attention for its ability to effectively infer tasks from contextual information. Notably, large language models such as GPT-3~\citep{brown2020language}, have shown remarkable abilities in tackling various tasks like text completion and code generation merely via language-based prompts. It also has been extended to the Reinforcement Learning (RL) area. \citet{laskin2022context} proposed the Algorithm Distillation (AD) algorithm, demonstrating in-context reinforcement learning by sequentially modeling offline data with an imitation loss. Later, \citet{lee2023supervised} developed the Decision-Pretrained Transformer (DPT), a transformer model pre-trained through supervised learning to predict optimal actions from an in-context dataset of interactions. 

This in-context learning ability aligns perfectly with our game-solving problem, as we aim for a model to effectively exploit any unknown opponent based on these online interactions with the opponent. 
For this reason, we make the first attempt to study the proposed game-solving problem in extensive-form games through in-context learning. 
In this paper, we propose a novel framework, In-Context Exploiter (ICE), designed to train a \textbf{\emph{single}} model that can \textbf{\emph{act as any player}} in the game to \textbf{\emph{adaptively exploit}} the opponents \textbf{\emph{without updating the parameters}}. A comparison of the traditional equilibrium finding framework with our ICE approach is depicted in Fig.\ref{fig:1}. Essentially, the ICE method can self-improve through in-context learning when against any opponent, unlike the NE strategy, which tends to adopt the safest, but not necessarily optimal, behavior. 

The ICE algorithm comprises three phases: i) generating opponent strategies through distinct approaches to ensure diversity; ii) gathering interactive history data using a reinforcement learning algorithm to maximize their utility against these opponent strategies; iii) training a transformer model on the history data within a curriculum learning framework to enhance the effectiveness and stability of the training process. In summary, our contributions are threefold: i) we make the first attempt to investigate a new game-solving problem through in-context learning; ii) we propose a novel algorithm, ICE algorithm, which possesses the capability of generalizability; iii) we conduct extensive experiments to demonstrate that ICE algorithm can effectively exploit any unknown opponent without updating parameters. 


\section{Background and Related Work}
\textbf{Imperfect-Information Extensive-Form Games.} An imperfect-information extensive-form game (EFG) can be represented by a tuple $(N, H, A, P, \mathcal{I}, u)$ \citep{shoham2008multiagent}. $N$ represents the set of players, i.e., $N=\{1,...,n\}$ and $H$ represents the set of histories which is the past action sequence. In particular, when the game starts, the history is an empty sequence $\emptyset$, representing the root node of the game tree. Additionally, every prefix of any sequence within $H$ is also included in $H$. There is a set of special histories, called terminal histories, which are sequences that end in the leaf nodes of the game tree. $Z$ is used to represent the set of terminal histories which is a subset of $H$, i.e., $Z \subset H$. $A(h) = \{a: (h, a)\in H\}$ represents the set of available actions at any non-terminal history $h \in H \setminus Z$. $P$ represents the player function which maps each non-terminal history to a player, i.e., $P(h) \mapsto N \cup \{c\}$ in which $c$ denotes chance player, representing these stochastic events beyond players' control. The information set, represented by $\mathcal{I}_i$, forms a partition over the set of histories where $i$ takes action, such that player $i \in N$ cannot distinguish these histories within the same information set $I_i$. Therefore, each information set $I_i \in \mathcal{I}_i$ corresponds to one decision-making point for player $i$ which means that $P(h_1)=P(h_2)$ and $A(h_1)=A(h_2)$ for any $h_1, h_2 \in I_i$. For convenience, we can employ $A(I_i)$ and $P(I_i)$ to denote $A(h)$ and $P(h)$ for any history $h$ within $I_i$. $u_i$ represents the utility function of player $i$ that maps every terminal history to real numbers, i.e., $u_i: Z \mapsto \mathbb{R}$. 

The behavior strategy for player $i$, denoted by $\sigma_i$, is a function that maps every information set $I_i$ to a probability distribution over the available action $A(I_i)$. The set of strategies for player $i$ is denoted by $\Sigma_i$, i.e., $\sigma_i \in \Sigma_i$. Given a strategy profile $\sigma=(\sigma_1, \sigma_2, ..., \sigma_n)$, the expected value to player $i$ is the sum of the expected payoff of these resulting terminal nodes, i.e, $u_i(\sigma)=\sum_{z\in Z}\pi^{\sigma}(z)u_i(z)$. $\pi^{\sigma}(z)=\prod_{i\in N \cup \{c\}}\pi_i^{\sigma}(z)$ is the reaching probability of terminal history $z$ and $\pi_i^{\sigma}(z)$ is the contribution of player $i$ to reach the terminal history $z$. 
The common solution concept for the imperfect-information extensive-form game is Nash equilibrium (NE) \citep{nash1950equilibrium}, defined as a strategy profile such that no player can increase their expected utility by unilaterally switching to a different strategy. Formally, a strategy profile $\sigma^*$ forms an NE if it satisfies $u_i(\sigma^*)=\max_{\sigma_i'\in \Sigma_i}u_i(\sigma_i', \sigma_{-i}^*), \forall i \in N$, where $\sigma_{-i}^*$ refers to all the strategies in $\sigma$ except for $\sigma_i$. The NE strategy is the safest and most stable strategy. However, it may not be the optimal strategy in many cases as we described in Sec.\ref{example}. In this paper, we aim to develop a model that can always effectively exploit different opponents to increase its utility.     

\textbf{In-Context Learning (ICL).} ICL is the ability to infer tasks and adapt strategies based on contextual information. A seminal example in this domain is the GPT series by OpenAI, particularly GPT-3 \citep{brown2020language}, which has demonstrated remarkable flexibility in handling a variety of tasks through language prompts alone. Recent advancements have also seen the integration of in-context learning into reinforcement learning domains. \citet{laskin2022context} introduced Algorithm Distillation (AD), a novel method that employs sequential modeling of offline data with an imitation loss for in-context reinforcement learning. This method has shown promising results in enhancing the adaptability of RL agents to a variety of environments. Afterward, \citet{lee2023supervised} proposed the Decision-Pretrained Transformer (DPT), a model pre-trained through supervised learning. This transformer model also exhibits its ability to solve a range of RL problems via in-context learning. These developments underscore the growing importance of in-context learning. By enabling models to intuitively adapt to new tasks based on contextual cues, in-context learning represents a significant step towards more flexible and general-purpose AI systems. Our research draws inspiration from these innovations, seeking to further explore and expand the capabilities of in-context learning in exploiting different opponents under extensive-form game settings.

\textbf{Curriculum Learning (CL).} CL has seen considerable exploration and application in recent years. This approach, inspired by the way humans learn, involves gradually increasing the complexity of learning tasks, thereby enhancing learning efficiency. \citet{bengio2009curriculum} were among the first to formalize the idea of curriculum learning in the context of machine learning. They demonstrated that starting with easier examples and progressively increasing difficulty could significantly improve the rate of convergence in training deep neural networks. This foundational work paved the way for a myriad of applications in various domains, such as natural language processing (NLP) \citep{zhang2018empirical,xu2020curriculum} and computer vision (CV) \citep{weinshall2018curriculum, hacohen2019power}. More recently, curriculum learning has been integrated into reinforcement learning. \citet{narvekar2020curriculum} provided a comprehensive overview of CL in RL, discussing how curricula can be designed to speed up the learning process by teaching agents simpler tasks before introducing more complex ones. Our problem also involves competing with different levels of opponents, thereby applying CL to our game-solving scenarios. By progressively increasing the complexity of game scenarios and opponent strategies, we aim to enhance the learning efficiency and adaptability ability of our model.

\section{Problem Statement}
\label{example}
We start by presenting our motivation with a well-known game of rock-paper-scissors, leading us to define our research problem based on the insights from this example. 

\begin{wraptable}{r}{0.5\textwidth}
\centering
\vspace{-18pt}
\caption{Rock-Paper-Scissors}\label{tab:rps}
\vspace{-5pt}  
\begin{tabular}{cccc}\\\toprule  
& R & P & S \\\midrule
R & (0,0) & (-1, 1) & (1,-1)\\  \midrule
P &(1, -1) &(0,0) & (-1, 1)\\  \midrule
S & (-1, 1) &(1, -1) & (0,0)\\  \bottomrule
\end{tabular}
\end{wraptable}
\textbf{NE is Safe, but not Optimal against Non-NE Opponent.} Given a rock-paper-scissors (RPS) game where the only NE is each player plays the three actions with the same probability, i.e., $1/3$, and the expected utility of each player is $0$. However, when the opponent is not rational, playing the NE strategy is not always preferred. Suppose that an opponent always plays rock, the expected utility of playing NE strategy against him is still $0$, while the expected utility of always playing paper is $1$. Therefore, playing the NE strategy would be a safe option but is not optimal in even two-player zero-sum symmetric games.

\textbf{Decision-making Problem.} The RPS game demonstrates that adherence to the Nash Equilibrium (NE) strategy, while safe, may not always yield optimal results, especially against non-rational players. This observation leads us to propose a novel game-solving problem: \emph{Can we learn \textbf{a model} which can exploit \textbf{any, even NE, opponent} to maximize their own utility?} In other words, can we build a model that can intelligently exploit different opponents to maximize their own utility across various strategic situations? In this paper, we start an initial exploration of investigating this game-solving problem and propose a framework, In-Context Exploiter (ICE), to train a model that can act as any player in the game to adaptively exploit the opponents without parameter updating, i.e., self-improvement entirely in context. 

\section{In-Context Exploiter}
In this section, we introduce our algorithm, In-Context Exploiter (ICE), specifically crafted to train a model to exploit any unknown opponent and increase its utility through its in-context learning ability. 
Fig.\ref{fig:method_overview} provides an overview of ICE methodology. As depicted, the ICE approach includes three primary stages: generating diverse opponent strategies, collecting interactive history through a reinforcement learning algorithm, and training a model within a curriculum learning framework. Each of these stages plays a critical role in ensuring the model’s adaptability and generalizability.

\subsection{Opponent Generation}
The key to developing a robust deep learning-based model that can effectively exploit various opponents is the creation of a comprehensive and representative set of opponent strategies for training. 
The diversity of these opponent strategies plays an important role, not only in enriching the training dataset but also in improving the model's capacity to generalize and adapt to exploit any new opponent through in-context learning. 
Our opponent generation method employs two approaches to ensure both diversity and representativeness: \textit{random generation} and \textit{learning-based generation}. 

\textbf{Random Generation.} Just as its name implies, random generation involves creating opponent strategies randomly. Specifically, for every opponent's possible information set, we randomly sample a policy for the information set to generate the opponent's strategy. This randomness ensures that the generated opponent strategy dataset is highly diverse and includes a variety of unpredictable opponent strategies, mimicking scenarios where opponents may act irrationally. 
\begin{figure}[t]
    \centering
\includegraphics[width=0.9\columnwidth]{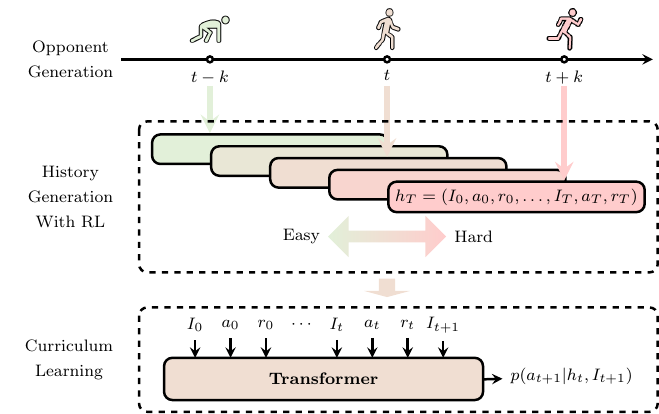}
\caption{Overview of ICE}
    \label{fig:method_overview}
\end{figure}

\textbf{Learning-based Generation.} The learning-based generation method focuses on collecting strategies opponents may employ while learning a game. To do this, we resort to equilibrium-finding algorithms, such as Counterfactual Regret Minimization (CFR) \citep{zinkevich2007regret} and Policy Space Response Oracle (PSRO) \citep{lanctot2017unified}. These algorithms are employed to solve a game, thereby collecting strategies that arise during their learning processes. As illustrated in Fig.\ref{fig:method_overview}, the opponents generated by this method progressively become more and more strategic. This progressive learning process ensures that our generated opponent strategy dataset includes a spectrum of skill levels, from naive to advanced. Such a range is essential for the training model to handle various strategic complexities.  

The combination of the aforementioned opponent generation methods is crucial in developing a comprehensive and diverse dataset. 
Such a dataset helps enhance the robustness and generalization of the model trained, equipping it to exploit opponents with varying levels of sophistication. 

\subsection{Interactive History Collection}
Given a diverse dataset consisting of opponent strategies, exploiting different opponent strategies can be regarded as a multi-task reinforcement learning problem. Then we can apply the multi-task pre-train and fine-tuning framework to solve this problem. 
However, the agent requires fine-tuning whenever meeting a new opponent. Considering the huge number of potential opponents, this framework is impractical for real-world tasks.
To mitigate these issues, we aim to train a model, equipping in-context learning ability, i.e., self-improvement without parameter updating. We borrow the idea from Algorithm Distillation (AD) \citep{laskin2022context} that distills RL algorithms into neural networks. 

In a word, we can utilize any suitable RL algorithm to exploit different opponents and then record these learning histories to distill a neural network model, equipping the in-context learning ability. In this section, we first introduce the collection of these historical data. Notably, each opponent's strategy corresponds to a unique learning task whose goal is to maximize their utility against the opponent. During the RL learning process, we record the interactive histories, capturing the interactions between the algorithm and the opponent. The recorded data comprises a sequence of information sets, actions taken, and the resulting utilities. The sequential data will be used for model training, as it provides rich, contextual insights into decision-making processes. In our experiments, we use proximal policy optimization (PPO) \citep{schulman2017proximal} to collect historical data, which is known to perform well across various tasks.

For each opponent, we can collect a piece of sequence data as shown in Fig.\ref{fig:method_overview}. This dataset forms the foundation for our transformer model, enabling our model to learn from the RL learning history. The use of RL algorithms in this context not only generates valuable training data but also contributes to the continuous improvement of our model's capabilities in competing against diverse opponents.  

\subsection{Curriculum Learning}
Note that after obtaining this learning historical dataset, we can train a model with the AD algorithm on these learning historical data directly. However, it may not be efficient or effective due to the high diversity of opponent strategies. The randomly ordered opponent strategies pose challenges in training, as the model must adapt to a wide range of behaviors and tactics, potentially leading to slower learning and reduced efficiency in the training process. To address this issue, we design a curriculum learning framework to enhance the stability and efficiency of the training process since the curriculum learning approach mimics the human learning process, where training begins with simpler tasks and gradually progresses to more complex challenges. 
\subsubsection{Curriculum Generation}
In this section, we outline the method for generating a curriculum. Since the curriculum learning mirrors the natural learning progression, we should generate the curriculum from simple tasks to difficult tasks. For the specific task of opponent exploitation, we recognize the difficulty of this task depends on the gap between the opponent strategy and the NE strategy since NE is the most difficult to exploit according to the definition of NE. It means that opponent strategies closer to NE are typically harder to exploit. Thus, we can generate the curriculum based on this understanding. 


\textbf{Natural Order of Learning-Generated Opponents.} Notably, the opponent's strategies generated by the learning-based generation method align naturally with the difficulty level, as they are sorted based on their proximity to NE. 
Thus, we can directly utilize this natural order to generate the curriculum for progressive training, starting from simpler strategies and moving towards more complex strategies. 

\textbf{Integration of Randomly-Generated Opponent.} Unfortunately, the opponent's strategies generated through the random generation method do not have the natural order. Of course, we can sort them according to the gap between these opponent strategies and the NE strategy. However, it would be time-consuming to compute these gaps since there are many opponent strategies. Fortunately, we find that most of the opponent's strategies generated through the random generation method are generally simple. Therefore, we can just intersperse these random opponents among the ordered opponents generated by the learning-based generation method to enhance the stability of the training process. 


The method for generating a curriculum is detailed in Algorithm \ref{alg:1}. It involves preserving the natural order of opponent strategies generated by the learning-based generation method. In this sequence, we strategically insert randomly generated opponent's strategies at regular intervals.


\begin{algorithm}[tb]   
\caption{Curriculum Generation}
\label{alg:1}
\begin{algorithmic}[1]
\STATE {\bfseries Input:} $S_l=[S_{l1}, ..., S_{ln}]$ opponent strategy dataset by learning-based generation method 
\STATE {\bfseries Input:} $S_r=[S_{r1}, ..., S_{rm}]$ opponent strategy dataset by random generation method
\STATE Initialize the gap $g$, the set $S_{order}=[]$;
\FOR{$i = 1$ to $m+n$}
\IF{$i \mod g = 0$}
\STATE $S_{order} = S_{order} \cup S_r[0]$ and $S_r$ delete $S_r[0]$;
\ELSE
\STATE $S_{order} = S_{order} \cup S_l[0]$ and $S_l$ delete $S_l[0]$;
\ENDIF
\IF{$|S_r|=0$ or $|S_l|=0$}
\STATE $S_{order}=S_{order} \cup S_l$ or $S_{order}=S_{order} \cup S_r$;
\STATE Early Break;
\ENDIF
\ENDFOR
\STATE {\bfseries Output:} The curriculum $S_{order}$
\end{algorithmic}
\end{algorithm}

\subsubsection{Transformer Training}
Before going into the details of the whole curriculum learning framework, we first introduce the training loss. Similar to the AD algorithm, we consider the training of the model as a sequence prediction problem. Therefore, any sequence model such as RNNs \citep{williams1989learning} can be used. In this paper, we adopt a causal transformer, renowned for its robustness in sequence modeling \citep{vaswani2017attention} and its adaptability to downstream tasks via in-context learning. 

For convenience, we represent the training data generated by the learning-based generation opponent strategies as $\mathcal{D}_l=\{\mathcal{D}_{l1}, ..., \mathcal{D}_{ln}\}$, and the training data from the random generation opponent strategies as $\mathcal{D}_r=\{\mathcal{D}_{r1}, ..., \mathcal{D}_{rm}\}$. These training data include the entire learning history for each task, capturing the sequence of interactions - information sets ($I_t$), actions ($a_t$), and rewards ($r_t$) - encountered when applying an RL algorithm. Formally, 
$\mathcal{D}_i = (I_0^{(i)}, a_0^{(i)}, r_0^{(i)},..., I_T^{(i)}, a_T^{(i)}, r_T^{(i)}), \mathcal{D}_i \in \mathcal{D}_r \cup \mathcal{D}_l.$

The essence of our approach lies in distilling the behaviors learned by the RL algorithms into the transformer. The transformer, denoted as $M_{\theta}$ with parameters $\theta$, is trained to map long histories (context) to probabilities over actions. To train the transformer, we adopt the negative log-likelihood (NLL) loss as the loss function, which can be formulated as:
\begin{align}
\mathcal{L}(\theta)=-\sum_{i=1}^{m+n}\sum_{t=1}^{T-1} \log M_{\theta}(A=a_t^{(i)}|h_{t-1}^{(i)}, I_{t}^{(i)})
\label{loss_function}
\end{align}
where $h_t$ represents the historical context up to $t$, defined as
$h_t = (I_0^{(i)}, a_0^{(i)}, r_0^{(i)},..., I_{t}^{(i)}, a_{t}^{(i)}, r_{t}^{(i)}) = (I_{\leq t}, a_{\leq t}, r_{\leq t}).$
This historical context contains the cumulative experience, which is critical for learning the in-context learning ability. 

During training, the transformer iteratively processes this historical data, learning to predict the next best action based on past sequences. Through this training process, the transformer model becomes adept at synthesizing long histories into optimal strategies, enhancing its generalizability capability to effectively exploit any unknown opponent.

\subsubsection{Step-by-step Training}
After generating a curriculum and defining the training loss for the transformer, we introduce how to train the transformer step-by-step according to the generated curriculum. A critical problem of training is catastrophic forgetting, especially when training on many tasks. Before into the training details, we introduce our method for mitigating this issue. 

\textbf{Preventing Catastrophic Forgetting.} To mitigate catastrophic forgetting, our training approach incorporates periodic reviews and retraining on previously tackled tasks. This continual learning approach ensures that the model retains its proficiency in earlier learned tasks while acquiring new capabilities. The rate of revising previous tasks is carefully calibrated to balance the retention of old knowledge with the acquisition of new skills, maintaining a comprehensive understanding across a spectrum of opponent strategies.  

\textbf{Training Process.} The whole curriculum training framework is depicted in Algorithm \ref{alg:adgs}. The curriculum $S_{order}$ and corresponding interactive history dataset $\mathcal{D}$ are taken as inputs. A transformer model $M_{\theta}$ is instantiated with parameters $\theta$ and a previous rate $\sigma$ is defined to control the blend of new and prior tasks to prevent catastrophic forgetting.
\begin{algorithm}[tb]  
\caption{Curriculum Training Framework}
   \label{alg:adgs}
\begin{algorithmic}[1]
   \STATE {\bfseries Input:} The curriculum $S_{order}=[S_{1}, ..., S_{N}]$
\STATE {\bfseries Input:} $\{\mathcal{D}_{1}, ..., \mathcal{D}_{N}\}$ corresponding dataset of tasks
   \STATE Initialize transformer $M_{\theta}$, previous rate $\sigma$ and $D=\{\}$;
\FOR{iteration $t=1$ to $T$}
    \FOR{train episode $p=1$ to $M$}
    \IF{$|D| < N$}
    \STATE \textbf{if} random number $> \sigma$ \textbf{then} $S_c = S_t$;
    \STATE \textbf{else} Sample one task $S_c$ from $D$; \textbf{end if}
    \ELSE
    \STATE Sample one task $S_c$ from $D$;
    \ENDIF
    \STATE Sample training data from dataset $\mathcal{D}_{S_c}$;
    \STATE Train $M_{\theta}$ on training batch according to Eq.(\ref{loss_function});
    \ENDFOR
    \STATE \textbf{if} $|D| < N$ \textbf{then} $D = D \cup \{S_t\}$; \textbf{end if};
\ENDFOR
\STATE {\bfseries Output:} The transformer $M_{\theta}$
\end{algorithmic}
\end{algorithm}

As the training iterations progress from $t=1$ to $T$, the algorithm systematically selects the current train task $S_t$ from the curriculum $S_{order}$. To avoid catastrophic forgetting, the training regimen incorporates a review mechanism where, upon training on a new task, the algorithm revisits previous tasks with a frequency determined by the rate $\sigma$ (Lines 9-13). This review strategy is crucial for maintaining and reinforcing the knowledge previously gained, thereby enhancing the model's ability for generalization across various scenarios. When all tasks have been trained, the algorithm shifts to a phase that consolidates learning by training across the entire task spectrum, i.e., randomly sampling a task from the trained task set (Line 15). After determining the training task, we perform the training on the corresponding dataset of the task $\mathcal{D}_{S_c}$ according to Eq.(\ref{loss_function}) (Lines 17-18). 









\section{Empirical Evaluation}
To assess the effectiveness of our ICE algorithm, we conduct comprehensive experiments on several popular extensive-form games. We begin by detailing our experimental setting. Then we provide a detailed analysis of the results, which are structured around answering several key research questions. 
\subsection{Experimental Setting}
\textbf{Experimental Subject.} For our experiments, we selected a range of poker games as test subjects, including two-player and three-player versions of Kuhn Poker, Leduc Poker, and Goofspiel with five cards. These games serve as diverse platforms to evaluate our algorithm's performance.

\textbf{Evaluation Testbed.}
To rigorously assess our algorithm, we constructed three distinct types of testbeds by randomly sampling opponents: \textit{in-distribution}, \textit{out-of-distribution}, and \textit{NE opponent}. For the in-distribution testbed, we selected approximately 30 opponent tasks from the task dataset utilized during training. For the out-of-distribution testbed, we randomly sampled 20 opponent strategies to create a diverse set of test tasks. Finally, for the NE opponent testbed, we specifically configured the opponent's strategy to align with the Nash Equilibrium, thereby forming test tasks that directly reflect NE strategies. This multifaceted testing approach allows us to thoroughly evaluate the adaptability and effectiveness of our algorithm in various strategic scenarios.

\begin{figure}[ht]
    \centering
    \subfigure[2-player Kuhn poker (left--player 1, right--player 2)]{
    \includegraphics[width=0.45\columnwidth]{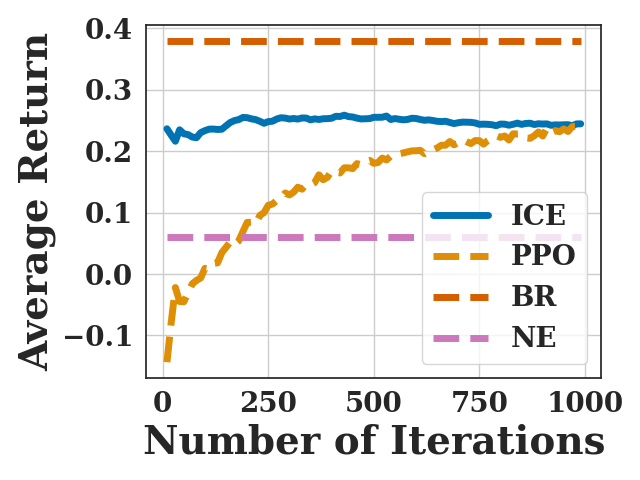}
\includegraphics[width=0.45\columnwidth]{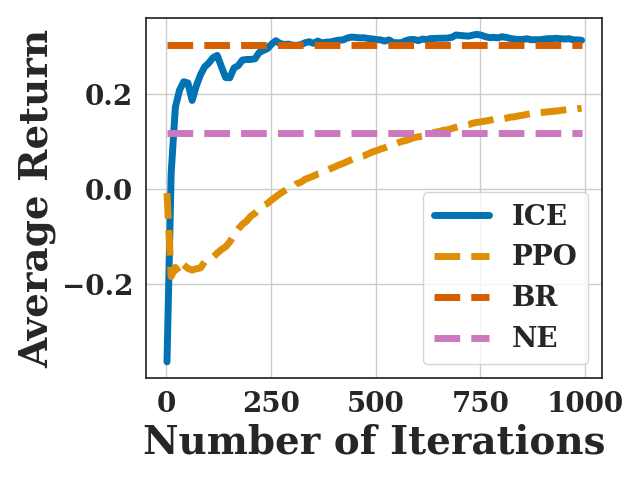}
     }
 \subfigure[3-player Kuhn poker (left--player 2, right--player 3)]{
     \includegraphics[width=0.45\columnwidth]{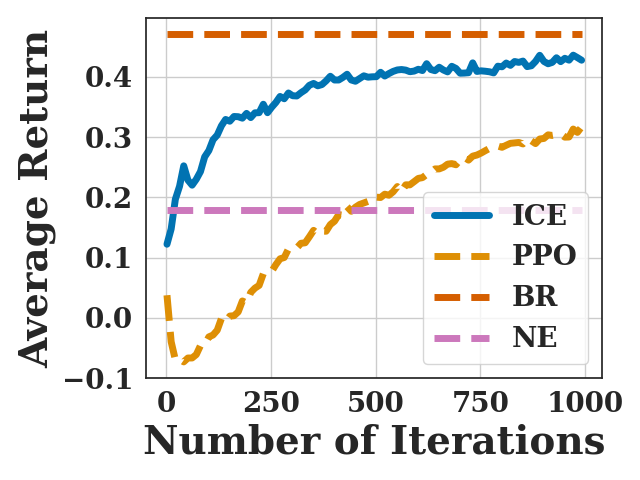} \includegraphics[width=0.45\columnwidth]{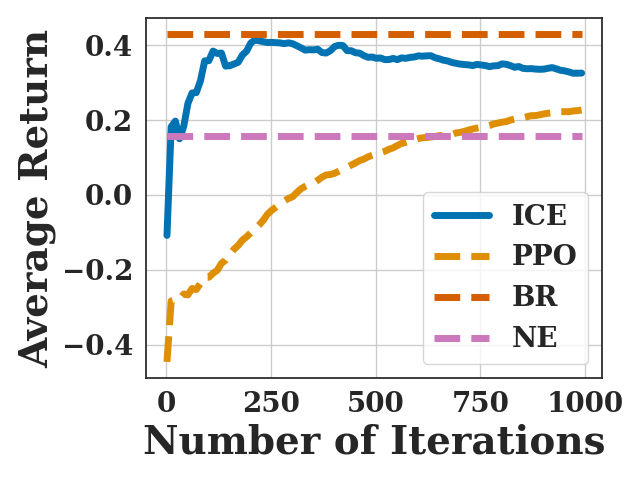}
     }
\caption{In-distribution results when acting as any player}
\label{fig:any_player}
\end{figure}

\begin{figure}[ht]
    \centering
    \subfigure[In-distribution results of three 2-player games (act as player 2)]{
    \includegraphics[width=0.32\columnwidth]{icml2024/figures/in_distribution/kuhn_poker_2_player/player-id_1_env-id_1.png}
\includegraphics[width=0.32\columnwidth]{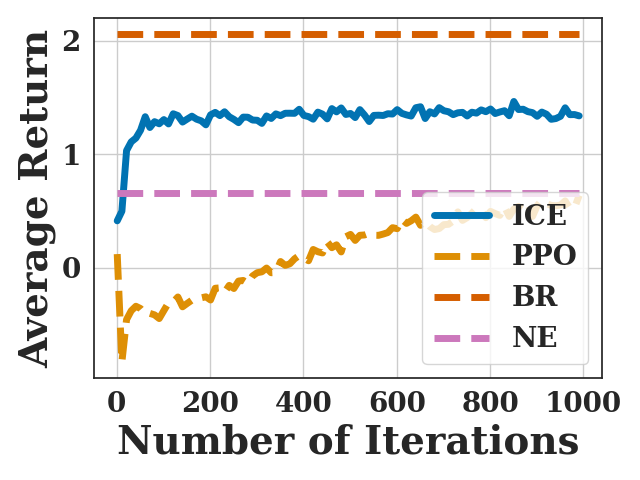}
\includegraphics[width=0.32\columnwidth]{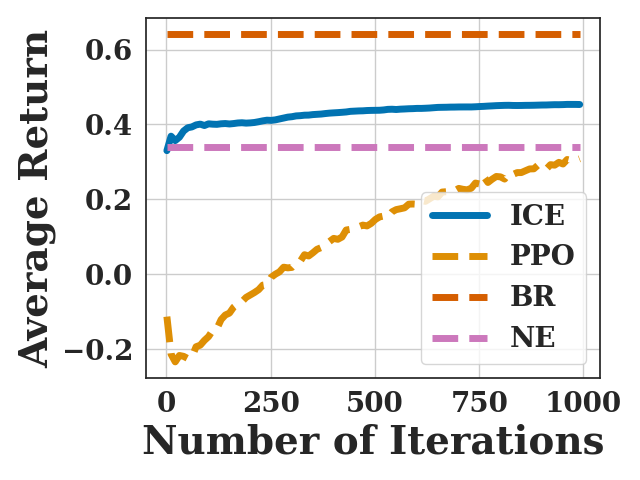}
     }
 \subfigure[Out-of-distribution results of three 2-player games (act as player 1)]{
     \includegraphics[width=0.32\columnwidth]{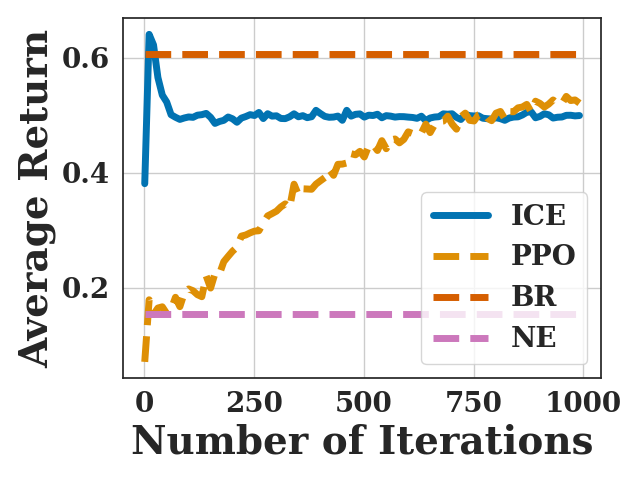} \includegraphics[width=0.32\columnwidth]{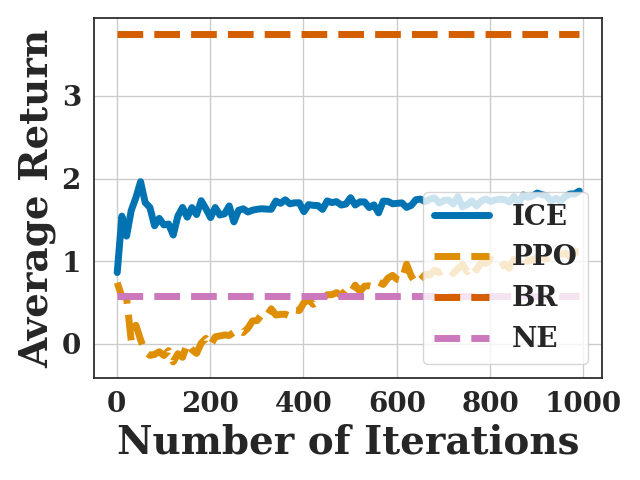}
     \includegraphics[width=0.32\columnwidth]{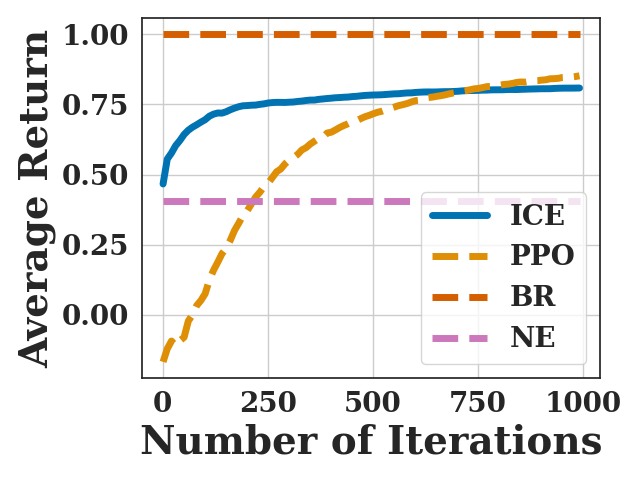}
     }
\caption{Results against various opponents (left-Kuhn, middle-Leduc, right-Goofspiel)}
\label{fig:various_opponent}
\end{figure}

\begin{figure}[ht]
    \centering
    \subfigure[2-player Kuhn poker]{
    \includegraphics[width=0.45\columnwidth]{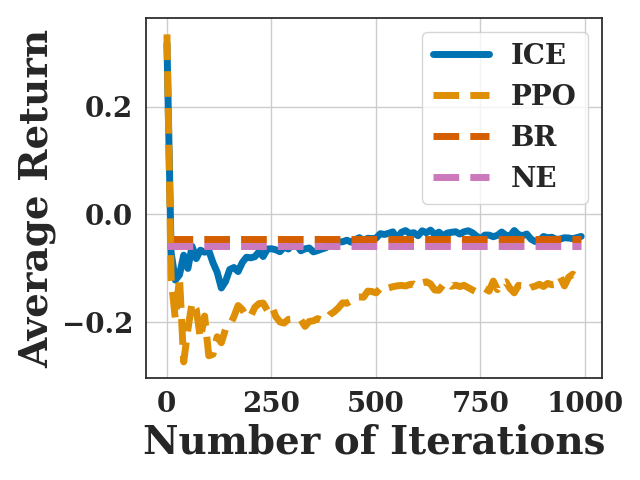}
     }
 \subfigure[2-player Leduc poker]{
     \includegraphics[width=0.45\columnwidth]{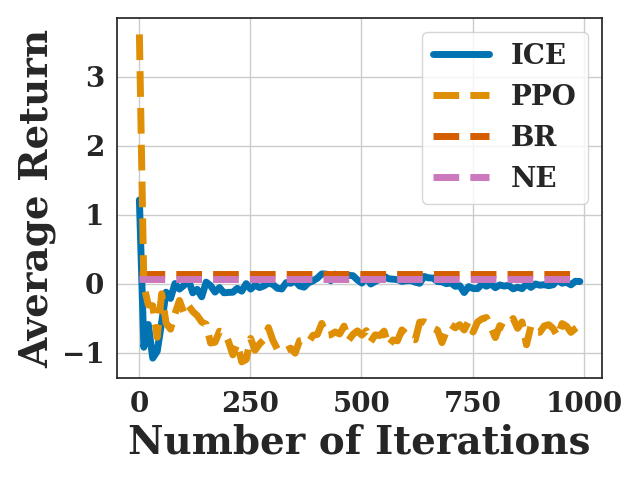}      }
\caption{Results against NE opponent}
\label{fig:ne_opponent}
\end{figure}

\textbf{Baselines.} We first choose two widely used strategies, the Best Response (BR) strategy and the Nash Equilibrium (NE) strategy, as baselines. Notably, BR is a theoretically optimal approach in an online setting since it is tailored against a known opponent's strategy. In the online setting, we include the BR strategy as a theoretical upper-bound benchmark. We select the NE strategy as a baseline since it is the most conservative strategy against any opponent. Since this is an online setting, we also select one online learning algorithm, proximal policy optimization (PPO) \citep{schulman2017proximal} algorithm as baselines. In addition to this, we also include multi-task pre-training with fine-tuning framework as one baseline since we can consider exploiting different opponents as multi-task learning. 
Since the BR and NE strategies are fixed once the opponent is given, we directly simulate these strategies against the opponent's strategy to evaluate their performance. 
For the PPO algorithm, multi-task pre-training with fine-tuning, and our ICE algorithm,  we conduct evaluations under a limited number of online interactions with the opponent. This limitation is deliberate, as our goal is to assess the capability of these algorithms to quickly and effectively exploit an unknown opponent. 

\subsection{Experimental Results}
To better demonstrate our results, we present our results by answering the following research questions (RQs). 

\textbf{RQ1:} \textit{Can the ICE algorithm act as any player in the game?} 
We claim that our ICE algorithm is capable of training a model to perform as any player in the game. To substantiate this claim, we conduct evaluations by positioning the model trained by the ICE algorithm in various player roles within the game. 
Fig.\ref{fig:any_player} presents the results for both two-player and three-player Kuhn poker, assessed using the in-distribution testbed. 
Our experimental results reveal that ICE outperforms both the NE strategy and the PPO algorithm when acting as any player of the game, whether in two-player or three-player games. Notably, ICE exhibits the capacity to self-improve and closely approximate the BR strategy, leveraging its in-context learning ability. These results show the effectiveness of our method in adjusting to various strategic roles. Consequently, we only show the results from the perspective of one player, as a representation of our algorithm's ability to adapt to and perform in any given role.

\textbf{RQ2:} \textit{Can our ICE adaptively exploit any opponent?}
To answer this question, we perform experiments on the three distinct testbeds we previously introduced, which simulate different opponents including NE opponents. This diverse range of testing environments is crucial to comprehensively evaluate the adaptability and effectiveness of our ICE algorithm in confronting any type of opponent. 
Fig.\ref{fig:various_opponent} and Fig.\ref{fig:ne_opponent} display the results of playing three two-player games against different opponents. We have also carried out experiments for three-player games, with those results included in the Appendix due to page constraints. From Fig.\ref{fig:various_opponent}, it is evident that our ICE algorithm effectively demonstrates its in-context learning capability. Within a limited number of interactions, our ICE surpasses both the NE strategy and the PPO algorithm. In simpler cases, the PPO algorithm may reach performance levels similar to that of our ICE algorithm. A key distinction, however, is that unlike PPO and other RL algorithms which require retraining from scratch for each new opponent, our ICE algorithm achieves this without any parameter updates.
In Figure \ref{fig:ne_opponent}, the average returns of NE and BR strategies are not exactly zero. This deviation arises from using an approximate NE strategy as the opponent. Additionally, the average returns for the NE and BR strategies are derived from simulations conducted with this approximate NE strategy. In this case, we observe that our ICE algorithm is capable of achieving results comparable to those of the NE and BR strategies, which means that our ICE can also adaptively exploit the NE opponent.    

\begin{figure}[t]
    \centering
    \subfigure[In-distribution result]{\includegraphics[width=0.45\columnwidth]{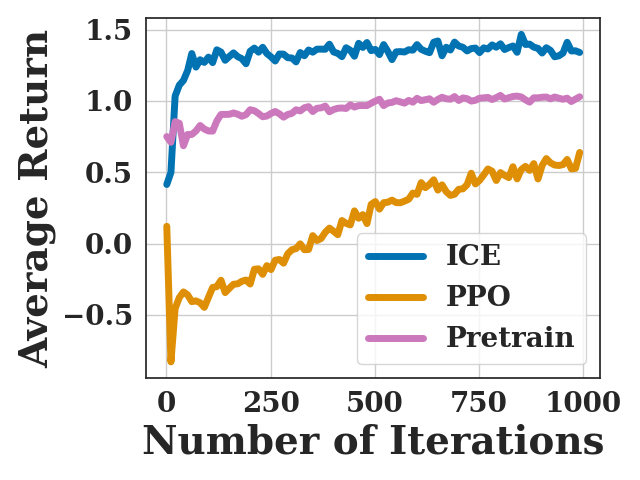}
     }
 \subfigure[Out-of-distribution result]{\includegraphics[width=0.45\columnwidth]{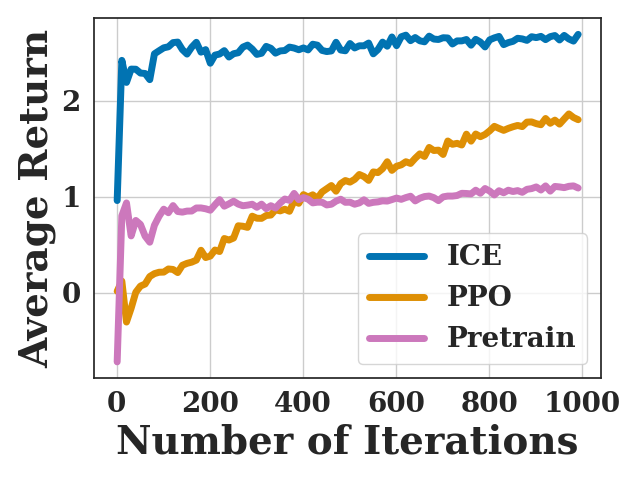}}
\caption{Comparison results with pre-training with fine-tuning}
\label{pre_train_compare}
\end{figure}

\textbf{RQ3:} \textit{How does our ICE algorithm perform compared with multi-task pre-training with fine-tuning framework?}
Recent work has shown that multi-task pre-training with fine-tuning on new tasks performs equally or better than meta-learning pre-training with meta adaptation in RL tasks~\citep{mandi2022effectiveness}. It indicates that pre-training with fine-tuning can quickly adapt to new tasks. In this paper, we compare this framework with our ICE algorithm. Firstly, we pre-train a model using tasks generated from opponents' strategies, the same as those used in the ICE algorithm. Then, we evaluate its performance by fine-tuning based on the interactions with the opponent. The results for a two-player Leduc poker game are depicted in Fig.\ref{pre_train_compare}. Our findings reveal that ICE outperforms pre-training with fine-tuning approach in both in-distribution and out-of-distribution testbeds. Notably, pre-training with fine-tuning performs even underperforms compared to the PPO algorithm in the out-of-distribution testbed. It might be attributed to the extensive potential opponent strategies, where pre-training cannot encompass all opponent types, leading to slower adaptation to new tasks. Additionally, the conflict in training direction for different player roles in zero-sum games could further hinder the effectiveness of pre-training with fine-tuning.    

\begin{figure}[t]
    \centering
    \subfigure[In-distribution result]{\includegraphics[width=0.45\columnwidth]{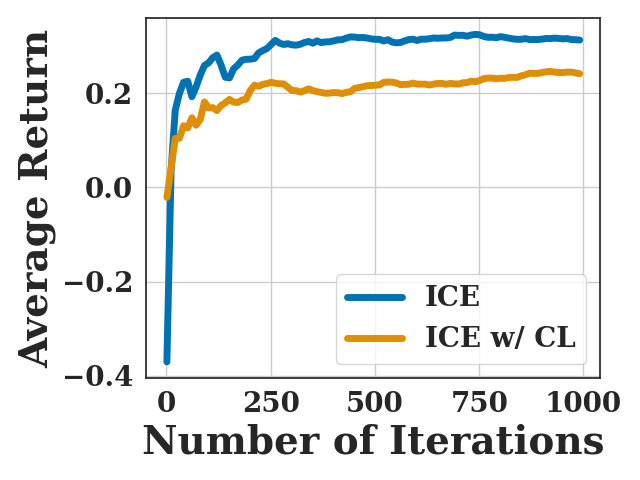}
     }
 \subfigure[Out-of-distribution result]{\includegraphics[width=0.45\columnwidth]{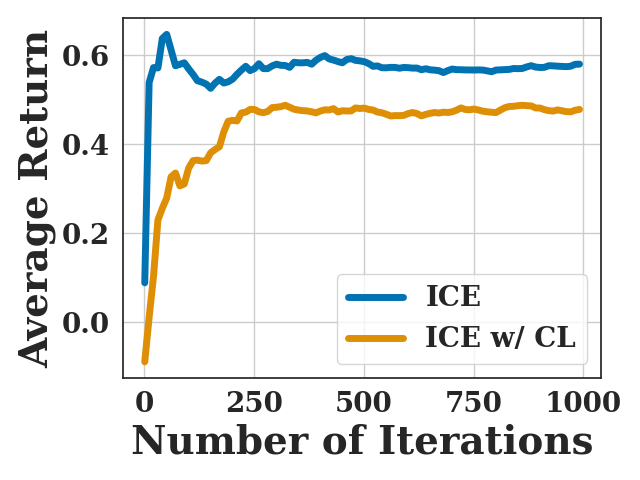}}
\caption{Results on 2-player Kuhn poker as player 2}
\label{CL_compare}
\end{figure}

\textbf{RQ4:} \textit{Can our curriculum learning (CL) framework enhance the performance?}
Our ICE algorithm incorporates a curriculum learning (CL) framework for training the transformer model. To explore the significance of CL, we conducted a comparative analysis by training the transformer model under a different setup, specifically without the CL framework. In this alternative scenario, the model is trained on a randomly ordered sequence of tasks. The comparative results for a two-player Kuhn poker game are presented in Fig.\ref{CL_compare}. We find that even in the absence of CL, the model retains its in-context learning ability when facing an unknown opponent. However, it is noteworthy that the ICE algorithm, when combined with the CL framework, consistently outperforms the version without CL in both in-distribution and out-of-distribution testbeds. This differential in performance underscores the significant contribution of the CL framework in boosting the effectiveness of our ICE algorithm. 

\textbf{RQ5:} \textit{Does the context length influence the performance?}
\begin{figure}[t]
\centering
    \subfigure[2-player Kuhn poker act as player 2]{\includegraphics[width=0.45\columnwidth]{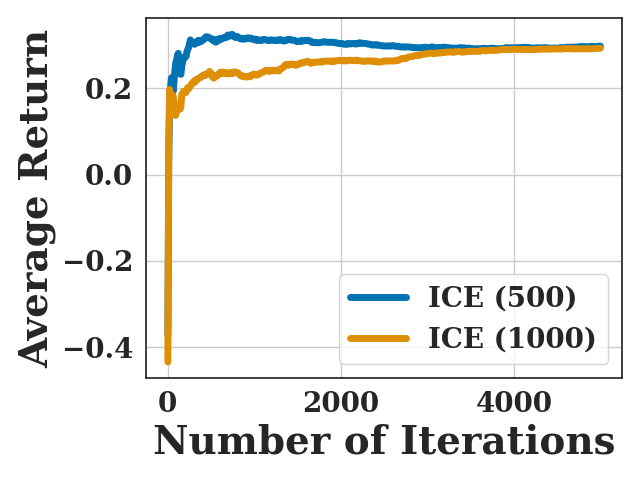}
     \includegraphics[width=0.45\columnwidth]{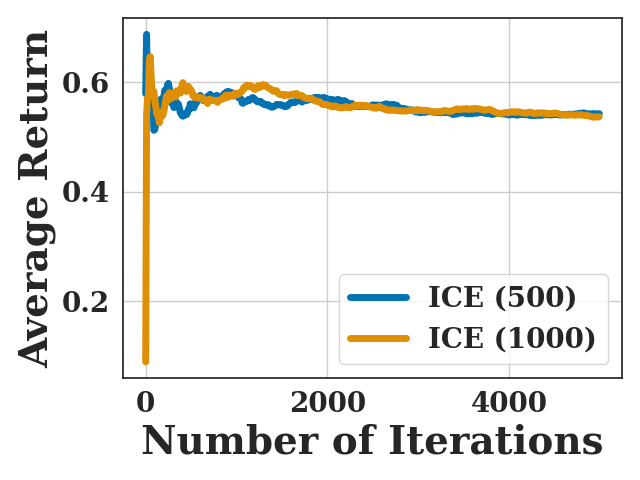}
     }
 \subfigure[3-player Goofspiel game act as player 2]{\includegraphics[width=0.45\columnwidth]{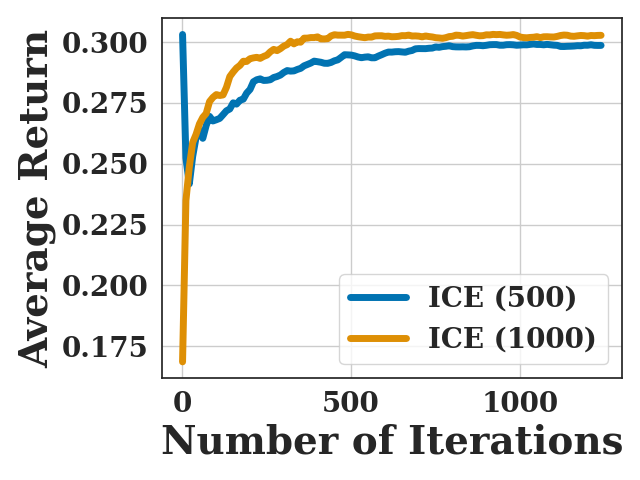}
 \includegraphics[width=0.45\columnwidth]{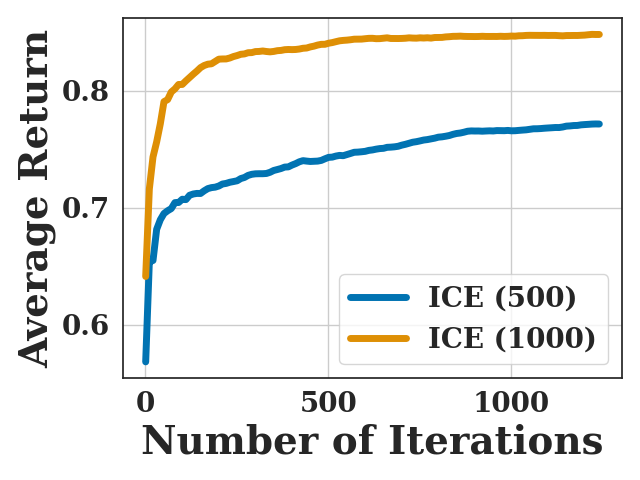}}
\caption{Results of different context lengths}
\label{context_length}
\end{figure}
Here, we examine how the pre-defined context length affects performance. To do this, we experimented with various context lengths in our game scenarios. The results for two-player Kuhn poker and three-player Goofspiel are illustrated in Fig.\ref{context_length}.
For Kuhn poker, the findings indicate that context length does not significantly impact performance. This might be due to the game's simplicity, where even a short context length suffices for effective in-context learning. Additionally, it's observed that at early stages, a larger context length may underperform compared to a shorter one, potentially requiring more interactions to fully utilize the extended context.
Conversely, in the Goofspiel game, the results suggest that a larger context length enhances performance. It implies that in intricate games, a large context, encompassing more interaction information, can significantly aid the decision-making process. The large context length provides a broader historical perspective, which is particularly beneficial in complex strategic environments where past interactions play a crucial role in future decisions.

\section{Conclusion}
In this paper, we investigate a pioneering game-solving problem: \textit{Can we learn a model that can exploit any opponent to maximize their utility?} To this end, we propose a framework, the In-Context Exploiter (ICE) algorithm, to train a single model that can act as any player in the game and adaptively exploit opponents through in-context learning. Our approach begins with generating opponent strategies to create diverse and representative tasks. We then apply an RL algorithm to solve these tasks, gathering the interactive history as training data. Subsequently, we develop a curriculum learning framework to effectively train a transformer model. Experimental results verify the effectiveness of the ICE algorithm in exploiting any unknown opponent and the model's ability to quickly adapt and optimize its strategies in various scenarios. The success of the ICE algorithm highlights the significant potential of in-context learning and this research not only answers our proposed game-solving problem affirmatively but also opens avenues for further exploration in the realm of strategic learning.


\bibliography{icml2024/example_paper}

\begin{thebibliography}{31}
\providecommand{\natexlab}[1]{#1}
\providecommand{\url}[1]{\texttt{#1}}
\expandafter\ifx\csname urlstyle\endcsname\relax
  \providecommand{\doi}[1]{doi: #1}\else
  \providecommand{\doi}{doi: \begingroup \urlstyle{rm}\Url}\fi

\bibitem[Albrecht \& Stone(2018)Albrecht and Stone]{albrecht2018autonomous}
Albrecht, S.~V. and Stone, P.
\newblock Autonomous agents modelling other agents: A comprehensive survey and open problems.
\newblock \emph{Artificial Intelligence}, 258:\penalty0 66--95, 2018.

\bibitem[Bard et~al.(2020)Bard, Foerster, Chandar, Burch, Lanctot, Song, Parisotto, Dumoulin, Moitra, Hughes, et~al.]{bard2020hanabi}
Bard, N., Foerster, J.~N., Chandar, S., Burch, N., Lanctot, M., Song, H.~F., Parisotto, E., Dumoulin, V., Moitra, S., Hughes, E., et~al.
\newblock The hanabi challenge: A new frontier for {AI} research.
\newblock \emph{Artificial Intelligence}, 280:\penalty0 103216, 2020.

\bibitem[Bengio et~al.(2009)Bengio, Louradour, Collobert, and Weston]{bengio2009curriculum}
Bengio, Y., Louradour, J., Collobert, R., and Weston, J.
\newblock Curriculum learning.
\newblock In \emph{Proceedings of the 26th Annual International Conference on Machine Learning}, pp.\  41--48, 2009.

\bibitem[Brown et~al.(2020)Brown, Mann, Ryder, Subbiah, Kaplan, Dhariwal, Neelakantan, Shyam, Sastry, Askell, et~al.]{brown2020language}
Brown, T.~B., Mann, B., Ryder, N., Subbiah, M., Kaplan, J., Dhariwal, P., Neelakantan, A., Shyam, P., Sastry, G., Askell, A., et~al.
\newblock Language models are few-shot learners.
\newblock In \emph{Proceedings of the 34th International Conference on Neural Information Processing Systems}, pp.\  1877--1901, 2020.

\bibitem[Finn et~al.(2017)Finn, Abbeel, and Levine]{finn2017model}
Finn, C., Abbeel, P., and Levine, S.
\newblock Model-agnostic meta-learning for fast adaptation of deep networks.
\newblock In \emph{International conference on machine learning}, pp.\  1126--1135, 2017.

\bibitem[Foerster et~al.(2017)Foerster, Chen, Al-Shedivat, Whiteson, Abbeel, and Mordatch]{foerster2017learning}
Foerster, J.~N., Chen, R.~Y., Al-Shedivat, M., Whiteson, S., Abbeel, P., and Mordatch, I.
\newblock Learning with opponent-learning awareness.
\newblock \emph{arXiv preprint arXiv:1709.04326}, 2017.

\bibitem[Hacohen \& Weinshall(2019)Hacohen and Weinshall]{hacohen2019power}
Hacohen, G. and Weinshall, D.
\newblock On the power of curriculum learning in training deep networks.
\newblock In \emph{Proceedings of the 36th Annual International Conference on Machine Learning}, pp.\  2535--2544, 2019.

\bibitem[He et~al.(2016)He, Boyd-Graber, Kwok, and Daum{\'e}~III]{he2016opponent}
He, H., Boyd-Graber, J., Kwok, K., and Daum{\'e}~III, H.
\newblock Opponent modeling in deep reinforcement learning.
\newblock In \emph{International Conference on Machine Learning}, pp.\  1804--1813, 2016.

\bibitem[Jain et~al.(2011)Jain, Korzhyk, Van{\v{e}}k, Conitzer, P{\v{e}}chou{\v{c}}ek, and Tambe]{jain2011double}
Jain, M., Korzhyk, D., Van{\v{e}}k, O., Conitzer, V., P{\v{e}}chou{\v{c}}ek, M., and Tambe, M.
\newblock A double oracle algorithm for zero-sum security games on graphs.
\newblock In \emph{Proceedings of the 10th International Conference on Autonomous Agents and Multi-Agent Systems}, pp.\  327--334, 2011.

\bibitem[Jain et~al.(2013)Jain, Conitzer, and Tambe]{jain2013security}
Jain, M., Conitzer, V., and Tambe, M.
\newblock Security scheduling for real-world networks.
\newblock In \emph{Proceedings of the 12rd International Conference on Autonomous Agents and Multi-Agent Systems}, pp.\  215--222, 2013.

\bibitem[Kurach et~al.(2020)Kurach, Raichuk, Sta{\'n}czyk, Zajac, Bachem, Espeholt, Riquelme, Vincent, Michalski, Bousquet, et~al.]{kurach2020google}
Kurach, K., Raichuk, A., Sta{\'n}czyk, P., Zajac, M., Bachem, O., Espeholt, L., Riquelme, C., Vincent, D., Michalski, M., Bousquet, O., et~al.
\newblock Google research football: A novel reinforcement learning environment.
\newblock In \emph{Proceedings of the AAAI conference on artificial intelligence}, volume~34, pp.\  4501--4510, 2020.

\bibitem[Lanctot et~al.(2017)Lanctot, Zambaldi, Gruslys, Lazaridou, Tuyls, P{\'e}rolat, Silver, and Graepel]{lanctot2017unified}
Lanctot, M., Zambaldi, V., Gruslys, A., Lazaridou, A., Tuyls, K., P{\'e}rolat, J., Silver, D., and Graepel, T.
\newblock A unified game-theoretic approach to multiagent reinforcement learning.
\newblock In \emph{Proceedings of the 31st International Conference on Neural Information Processing Systems}, pp.\  4193--4206, 2017.

\bibitem[Laskin et~al.(2022)Laskin, Wang, Oh, Parisotto, Spencer, Steigerwald, Strouse, Hansen, Filos, Brooks, et~al.]{laskin2022context}
Laskin, M., Wang, L., Oh, J., Parisotto, E., Spencer, S., Steigerwald, R., Strouse, D., Hansen, S., Filos, A., Brooks, E., et~al.
\newblock In-context reinforcement learning with algorithm distillation.
\newblock \emph{arXiv preprint arXiv:2210.14215}, 2022.

\bibitem[Lee et~al.(2023)Lee, Xie, Pacchiano, Chandak, Finn, Nachum, and Brunskill]{lee2023supervised}
Lee, J.~N., Xie, A., Pacchiano, A., Chandak, Y., Finn, C., Nachum, O., and Brunskill, E.
\newblock Supervised pretraining can learn in-context reinforcement learning.
\newblock \emph{arXiv preprint arXiv:2306.14892}, 2023.

\bibitem[Mandi et~al.(2022)Mandi, Abbeel, and James]{mandi2022effectiveness}
Mandi, Z., Abbeel, P., and James, S.
\newblock On the effectiveness of fine-tuning versus meta-reinforcement learning.
\newblock \emph{arXiv preprint arXiv:2206.03271}, 2022.

\bibitem[Morav{\v{c}}{\'\i}k et~al.(2017)Morav{\v{c}}{\'\i}k, Schmid, Burch, Lis{\`y}, Morrill, Bard, Davis, Waugh, Johanson, and Bowling]{moravvcik2017deepstack}
Morav{\v{c}}{\'\i}k, M., Schmid, M., Burch, N., Lis{\`y}, V., Morrill, D., Bard, N., Davis, T., Waugh, K., Johanson, M., and Bowling, M.
\newblock {DeepStack}: Expert-level artificial intelligence in heads-up no-limit poker.
\newblock \emph{Science}, 356\penalty0 (6337):\penalty0 508--513, 2017.

\bibitem[Narvekar et~al.(2020)Narvekar, Peng, Leonetti, Sinapov, Taylor, and Stone]{narvekar2020curriculum}
Narvekar, S., Peng, B., Leonetti, M., Sinapov, J., Taylor, M.~E., and Stone, P.
\newblock Curriculum learning for reinforcement learning domains: A framework and survey.
\newblock \emph{Journal of Machine Learning Research}, 21\penalty0 (181):\penalty0 1--50, 2020.

\bibitem[Nash(1950)]{nash1950equilibrium}
Nash, J.
\newblock Equilibrium points in n-person games.
\newblock \emph{Proceedings of the National Academy of Sciences of the United States of America}, 36\penalty0 (1):\penalty0 48--49, 1950.

\bibitem[Perolat et~al.(2022)Perolat, De~Vylder, Hennes, Tarassov, Strub, de~Boer, Muller, Connor, Burch, Anthony, et~al.]{perolat2022mastering}
Perolat, J., De~Vylder, B., Hennes, D., Tarassov, E., Strub, F., de~Boer, V., Muller, P., Connor, J.~T., Burch, N., Anthony, T., et~al.
\newblock Mastering the game of {Stratego} with model-free multiagent reinforcement learning.
\newblock \emph{Science}, 378\penalty0 (6623):\penalty0 990--996, 2022.

\bibitem[Schulman et~al.(2017)Schulman, Wolski, Dhariwal, Radford, and Klimov]{schulman2017proximal}
Schulman, J., Wolski, F., Dhariwal, P., Radford, A., and Klimov, O.
\newblock Proximal policy optimization algorithms.
\newblock \emph{arXiv preprint arXiv:1707.06347}, 2017.

\bibitem[Shalev-Shwartz et~al.(2012)]{shalev2012online}
Shalev-Shwartz, S. et~al.
\newblock Online learning and online convex optimization.
\newblock \emph{Foundations and Trends{\textregistered} in Machine Learning}, 4\penalty0 (2):\penalty0 107--194, 2012.

\bibitem[Shoham \& Leyton-Brown(2008)Shoham and Leyton-Brown]{shoham2008multiagent}
Shoham, Y. and Leyton-Brown, K.
\newblock \emph{Multiagent Systems: Algorithmic, Game-Theoretic, and Logical Foundations}.
\newblock Cambridge University Press, 2008.

\bibitem[Sumers et~al.(2023)Sumers, Yao, Narasimhan, and Griffiths]{sumers2023cognitive}
Sumers, T.~R., Yao, S., Narasimhan, K., and Griffiths, T.~L.
\newblock Cognitive architectures for language agents.
\newblock \emph{arXiv preprint arXiv:2309.02427}, 2023.

\bibitem[Tammelin et~al.(2015)Tammelin, Burch, Johanson, and Bowling]{tammelin2015solving}
Tammelin, O., Burch, N., Johanson, M., and Bowling, M.
\newblock Solving heads-up limit texas hold'em.
\newblock In \emph{Proceedings of the 24th International Conference on Artificial Intelligence}, pp.\  645--652, 2015.

\bibitem[Vaswani et~al.(2017)Vaswani, Shazeer, Parmar, Uszkoreit, Jones, Gomez, Kaiser, and Polosukhin]{vaswani2017attention}
Vaswani, A., Shazeer, N., Parmar, N., Uszkoreit, J., Jones, L., Gomez, A.~N., Kaiser, {\L}., and Polosukhin, I.
\newblock Attention is all you need.
\newblock In \emph{Proceedings of the 31st International Conference on Neural Information Processing Systems}, pp.\  6000--6010, 2017.

\bibitem[Weinshall et~al.(2018)Weinshall, Cohen, and Amir]{weinshall2018curriculum}
Weinshall, D., Cohen, G., and Amir, D.
\newblock Curriculum learning by transfer learning: Theory and experiments with deep networks.
\newblock In \emph{Proceedings of the 35th Annual International Conference on Machine Learning}, pp.\  5238--5246, 2018.

\bibitem[Williams \& Zipser(1989)Williams and Zipser]{williams1989learning}
Williams, R.~J. and Zipser, D.
\newblock A learning algorithm for continually running fully recurrent neural networks.
\newblock \emph{Neural computation}, 1\penalty0 (2):\penalty0 270--280, 1989.

\bibitem[Wu et~al.(2022)Wu, Li, Xu, Zang, An, and Xing]{wu2022l2e}
Wu, Z., Li, K., Xu, H., Zang, Y., An, B., and Xing, J.
\newblock L2e: Learning to exploit your opponent.
\newblock In \emph{2022 International Joint Conference on Neural Networks (IJCNN)}, pp.\  1--8, 2022.

\bibitem[Xu et~al.(2020)Xu, Zhang, Mao, Wang, Xie, and Zhang]{xu2020curriculum}
Xu, B., Zhang, L., Mao, Z., Wang, Q., Xie, H., and Zhang, Y.
\newblock Curriculum learning for natural language understanding.
\newblock In \emph{Proceedings of the 58th Annual Meeting of the Association for Computational Linguistics}, pp.\  6095--6104, 2020.

\bibitem[Zhang et~al.(2018)Zhang, Kumar, Khayrallah, Murray, Gwinnup, Martindale, McNamee, Duh, and Carpuat]{zhang2018empirical}
Zhang, X., Kumar, G., Khayrallah, H., Murray, K., Gwinnup, J., Martindale, M.~J., McNamee, P., Duh, K., and Carpuat, M.
\newblock An empirical exploration of curriculum learning for neural machine translation.
\newblock \emph{arXiv preprint arXiv:1811.00739}, 2018.

\bibitem[Zinkevich et~al.(2007)Zinkevich, Johanson, Bowling, and Piccione]{zinkevich2007regret}
Zinkevich, M., Johanson, M., Bowling, M., and Piccione, C.
\newblock Regret minimization in games with incomplete information.
\newblock In \emph{Proceedings of the 20th International Conference on Neural Information Processing Systems}, pp.\  1729--1736, 2007.

\end{thebibliography}
\bibliographystyle{icml2024/icml2024}


\appendix

\clearpage
\section{Discussion}
\subsection{Comparing ICE with Opponent Modeling}
ICE can be viewed as a method with \textbf{implicit} opponent modeling~\citep{he2016opponent,albrecht2018autonomous}, where the modeling of the opponent is implicitly encoded into the parameters of the model. There are several advantages of ICE over opponent modeling: i) ICE does not need an explicit model for the opponents, where the explicit model in the opponent modeling may restrict the generalizability of the methods, ii) ICE can exploit different opponents without changing the parameters, where the opponent modeling may need to fit the parameters of the opponent model during game play and then make the decision in response to the opponent. To summarize, ICE is simpler, more efficient, and more generalizable. ICE also has the disadvantage, i.e., the ability to model the opponents is largely determined by the length of the in-context. With longer in-context, ICE can model more opponents, while the model will also be larger and the training cost will be increased. We will discuss the methods to reduce the length of the in-context in the next section.
On the other hand, we can also introduce an explicit model for opponents into ICE, where the parameters of the opponent model can be fitted through in-context learning. The explicit opponent model can help us to understand the internal mechanism of ICE. 
\subsection{Comparing ICE with Online Learning, Multitask Learning, and Meta Learning}
ICE, as well as other in-context learning methods~\citep{laskin2022context,lee2023supervised}, is similar to online learning methods, e.g., no-regret learning~\citep{shalev2012online}. However, ICE does not change the parameters of the model during the game play with the opponents, which differs from online learning. We believe that online learning, especially no-regret learning, can be used to analyze the behaviors of ICE and in-context learning methods, which will be explored in future works. 
We also consider online learning methods as our baselines. We note that PPO is an online learning and on-policy method and PPO is scalable and widely used. Therefore, we include PPO as the baseline in our experiments. 

The training of ICE is also similar to multitask learning~\citep{mandi2022effectiveness}
and meta-learning~\citep{finn2017model}, where multi-task learning learns a policy for different tasks, and meta-learning enables the fast adaption of the learned policy on specific tasks. ICE also learns a policy for different tasks, where the model parameters are not changed, but the behaviors are changed during game play. In the experiment section, we choose the PPO method initialized with a pre-trained policy to benchmark multi-task and meta-learning methods, as shown in Figure~\ref{pre_train_compare}.

\subsection{Limitations and Future Works}
In this section, we provide a detailed discussion about the limitations of ICE and the future works. 
\paragraph{Dynamic Opponents.} The main objective of this work is to demonstrate the generalizability of in-context learning in solving extensive-form games, therefore we only consider the different static opponents, i.e., the opponents' policies are not changed during playing with our model. In future work, we will consider applying ICE against dynamic opponents, where the opponents can be rule-based agents, learning agents, or even in-context learning agents. The dynamic opponents will bring extra difficulties to ICE, including the enormous types of dynamic opponents, the instabilities when all players are changing their behaviors, and the difficulties for training. Therefore, novel methods are needed to make ICE robustly and safely exploit dynamic opponents without being exploited.

\paragraph{Reducing the In-Context Length.} As discussed above, the length of the in-context will significantly influence the ability of ICE, which is also demonstrated in our experiments. To handle the problems with more complicated dynamics and higher dimensions of the observations, the input dimensions of the model will grow drastically. Therefore, novel methods are required to reduce the in-context length of ICE. One possible option is that we can introduce the memory~\citep{sumers2023cognitive} to ICE, either internal or external, and we can query the relevant experiences from the memory to form the in-context. 

\paragraph{Generalizability of ICE.} In this work, we only consider the case where the model will play against different opponents of a game. A more challenging case for the generalizability of ICE is applying the model learned by ICE to different games and different opponents, which requires a unified representation of the observations and actions of different games. Future efforts will be dedicated to overcoming these obstacles, aiming to extend the applicability of ICE to different games, e.g., Stratego~\citep{perolat2022mastering}, and decision-making scenarios such as cooperative games, e.g., Handbi~\citep{bard2020hanabi},  and even mixed cooperative and competitive games, e.g., football~\citep{kurach2020google}.


\section{Implementation Details}
In this section, we provide the experimental details of our ICE algorithm from its three main stages. 

\textbf{Opponent Generation.} In this paper, we employ two methods, as introduced in the main paper, to generate a diverse range of opponent strategies. To implement the random generation method, we traverse through all the information sets of an opponent and assign a randomly generated strategy to each information set. This approach allows us to generate various opponents exhibiting random behaviors. To implement the learning-based generation method, we utilize a well-known algorithm, Counterfactual Regret Minimization (CFR) \citep{zinkevich2007regret},  as the equilibrium-finding algorithm. By applying CFR to solve the game, we record the average strategy for each player at each iteration. This process generates a series of opponent strategies that evolve from random to increasingly robust over time. These two methods collectively ensure that our dataset includes a wide spectrum of opponent strategies, ranging from entirely unpredictable to highly strategic. Such a comprehensive dataset is instrumental in training our model to adapt and respond effectively to various levels of opponent sophistication and strategy.

\textbf{Interactive History Collection.} 
It's important to recognize that when an opponent's strategy is known, the task of exploiting that opponent to maximize utility effectively becomes a reinforcement learning (RL) problem. Consequently, each distinct opponent strategy corresponds to a unique RL task. To collect interactive history data from our diverse opponent strategies for training purposes, we adopt the Proximal Policy Optimization (PPO) algorithm \citep{schulman2017proximal} to address each of these RL tasks. During this process, we systematically record the learning history of the PPO algorithm, specifically capturing the contents of the reply buffer used by PPO.

\textbf{Curriclum Learning.} The curriculum learning framework plays a pivotal role in effectively training the transformer model, with the core aspect being the design of the curriculum itself. In the main paper, we have thoroughly detailed the process of generating the curriculum and the overarching structure of the learning framework. This section will not delve into the specifics of the curriculum learning framework. However, it's important to emphasize that the strategic design of the curriculum is integral to the success of our model's training. The gradual escalation in complexity and the structured progression of tasks ensure that the model is not overwhelmed and can build its understanding and capabilities incrementally. This approach aligns with the principles of in-context learning, enabling the transformer to adapt and respond effectively to a wide range of strategic scenarios.

\textbf{Parameter Setting.} Here, we list the parameters used in the ICE algorithm for all games in Tab.\ref{parameter}. In this table, the previous rate $\sigma$ is used to control the blend of new and prior tasks to prevent catastrophic forgetting and the number of trains per task refers to the number of training for each selected task (i.e., $M$ in Algorithm \ref{alg:adgs}). 

\begin{table}[ht]
\centering
\caption{Parameter for ICE} \label{parameter}
\begin{tabular}{c|c|c|c|c|c|c}
\toprule  
Games & Kuhn & Leduc & Goofspiel &  Kuhn & Leduc & Goofspiel \\
\midrule
Number of Player & 2 & 2&2&3&3&3  \\
\midrule
Previous Rate $\sigma$ & 0.1 & 0.3 & 0.3 & 0.1 & 0.3 & 0.3\\  \midrule
Number of Train per Task $M$ & 10 & 10 & 10 & 10 & 30 & 30\\  \midrule
Context Length& 1000 & 1000 &1000&1000 &1000 &1000\\  \bottomrule
\end{tabular}
\end{table}

\section{Additional Experimental Results.}
In this section, we present further experimental results to substantiate the effectiveness of our ICE algorithm. While the main paper provided the performance of ICE in three two-player game scenarios evaluated across three distinct testbeds, here we extend our analysis to include results from experiments conducted on three different three-player games.
\begin{figure}[ht]
    \centering
    \subfigure[3-player Kuhn poker (left--player 1, middle--player 2, right--player 3)]{
    \includegraphics[width=0.32\columnwidth]{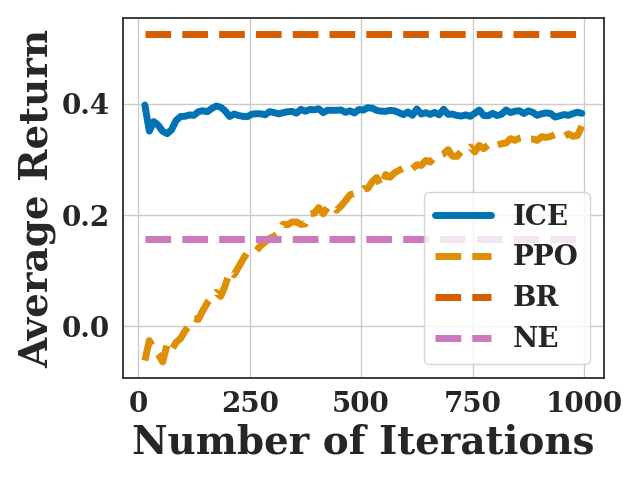}
\includegraphics[width=0.32\columnwidth]{icml2024/figures/in_distribution/kuhn_poker_3_player/player-id_1_env-id_2.png}
\includegraphics[width=0.32\columnwidth]{icml2024/figures/in_distribution/kuhn_poker_3_player/player-id_2_env-id_0.png}
     }
 \subfigure[3-player Leduc poker (left--player 1, middle--player 2, right--player 3)]{
     \includegraphics[width=0.32\columnwidth]{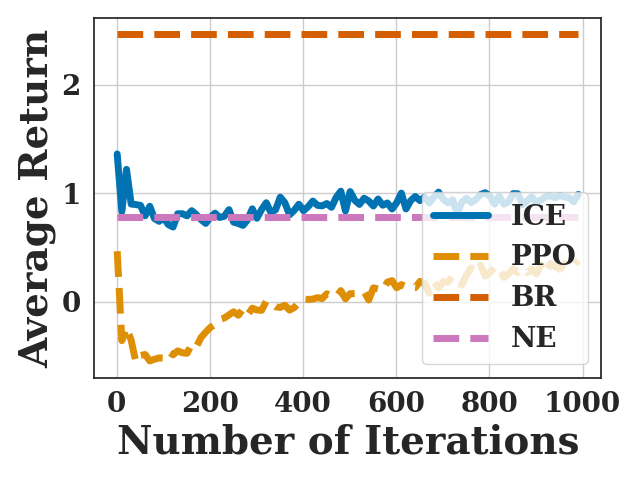} \includegraphics[width=0.32\columnwidth]{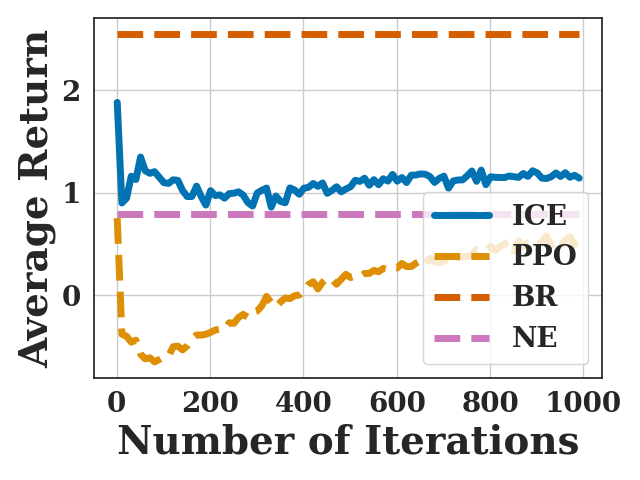}
     \includegraphics[width=0.32\columnwidth]{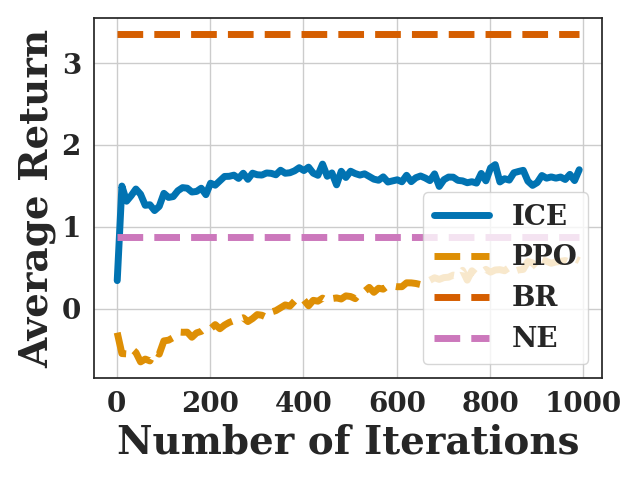}}
 \subfigure[3-player Goofspiel (left--player 1, middle--player 2, right--player 3)]{
     \includegraphics[width=0.32\columnwidth]{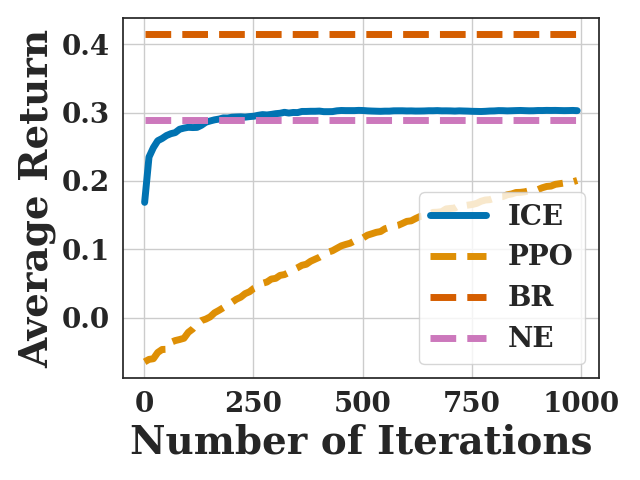} \includegraphics[width=0.32\columnwidth]{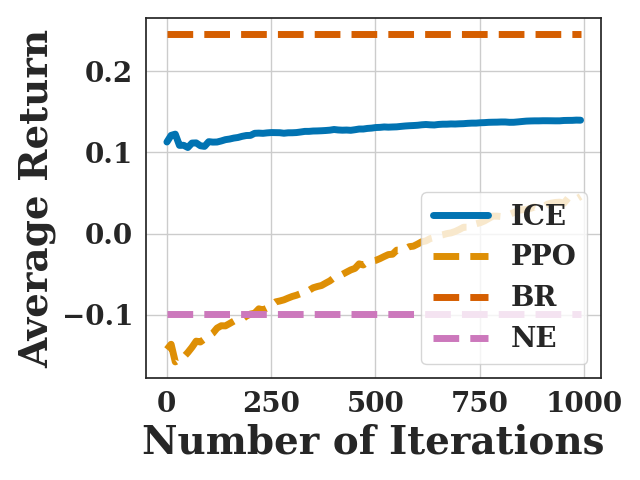}
     \includegraphics[width=0.32\columnwidth]{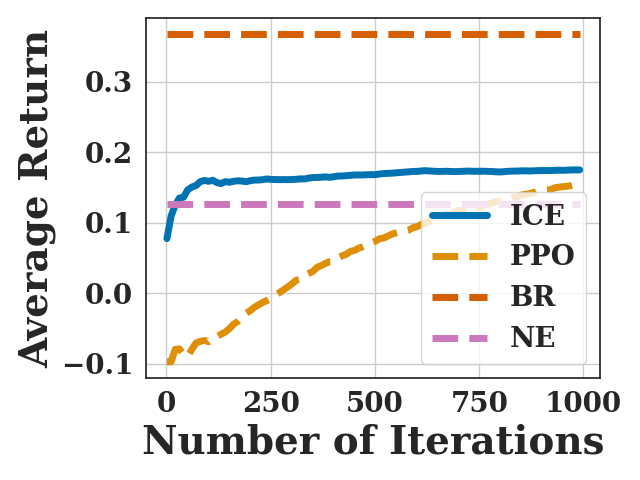}}
\caption{In-distribution results of three-player games}
\label{fig:3_player_in_distribution}
\end{figure}

Firstly, we present the results from the in-distribution testbed in Fig.\ref{fig:3_player_in_distribution}. In these three three-player games, it is evident that the model trained using the ICE algorithm successfully functions as any player in the game, demonstrating in-context learning ability with increasing iterations. The Best Response (BR) strategy, while theoretically the optimal approach since it is tailored against a known opponent's strategy, isn't practical in real-world scenarios where an opponent's strategy isn't known in advance. In our results, the BR strategy's performance is included merely as a theoretical benchmark. Notably, while the ICE-trained model doesn't achieve the theoretical optimal values of the BR strategy, it consistently surpasses both the NE strategy and the PPO algorithm. This observation is significant as it indicates that the ICE-trained model can exploit the opponents more effectively than the NE strategy, which is generally considered the most conservative approach. The ability of ICE to outperform in these multi-player game scenarios demonstrates its potential as a powerful tool for strategic decision-making in complex, real-world situations. 

\begin{figure}[t]
    \centering
    \subfigure[3-player Kuhn poker (left--player 1, middle--player 2, right--player 3)]{
    \includegraphics[width=0.32\columnwidth]{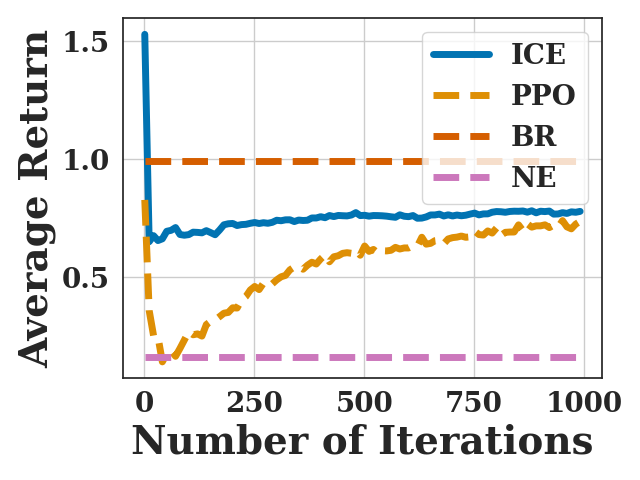}
\includegraphics[width=0.32\columnwidth]{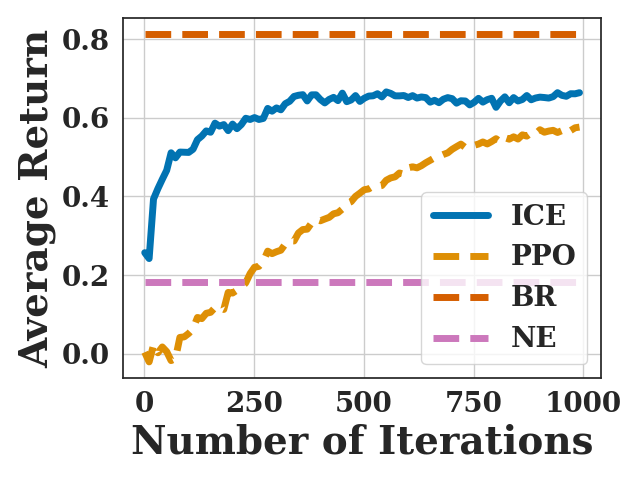}
\includegraphics[width=0.32\columnwidth]{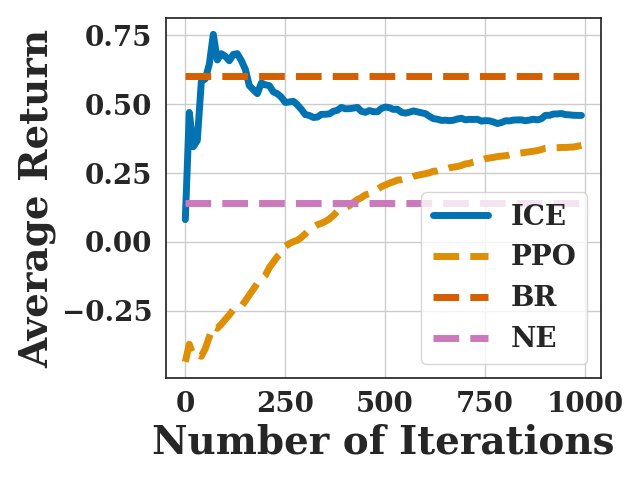}
     }
 \subfigure[3-player Leduc poker (left--player 1, middle--player 2, right--player 3)]{
     \includegraphics[width=0.32\columnwidth]{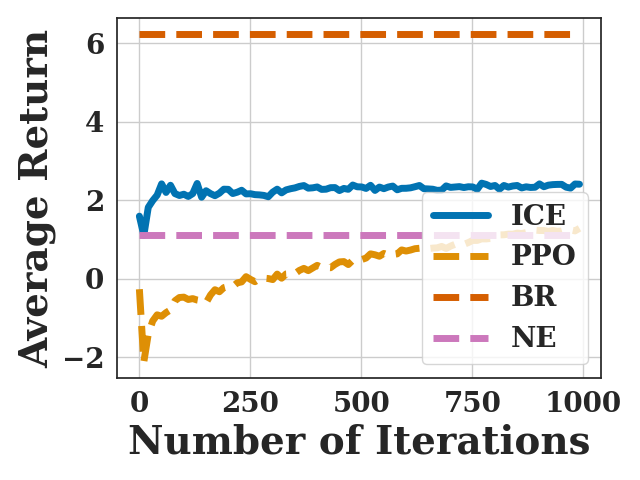} \includegraphics[width=0.32\columnwidth]{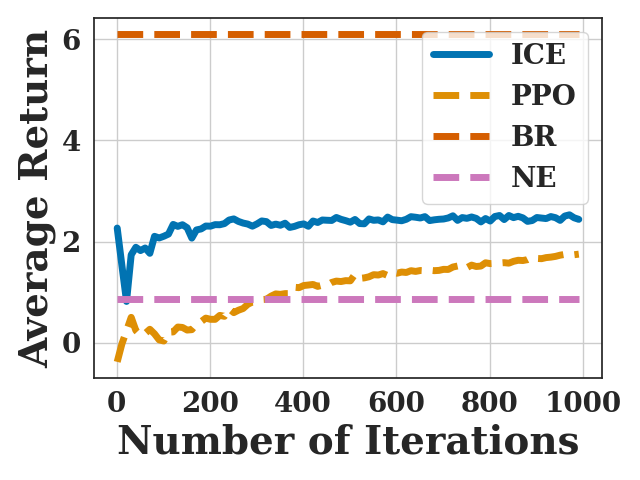}
     \includegraphics[width=0.32\columnwidth]{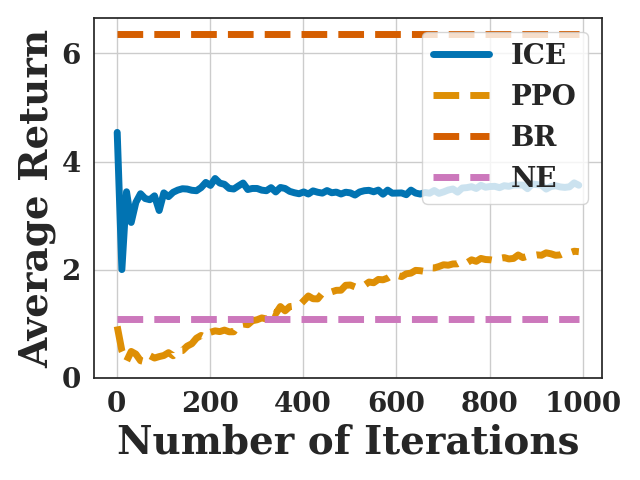}}
 \subfigure[3-player Goofspiel (left--player 1, middle--player 2, right--player 3)]{
     \includegraphics[width=0.32\columnwidth]{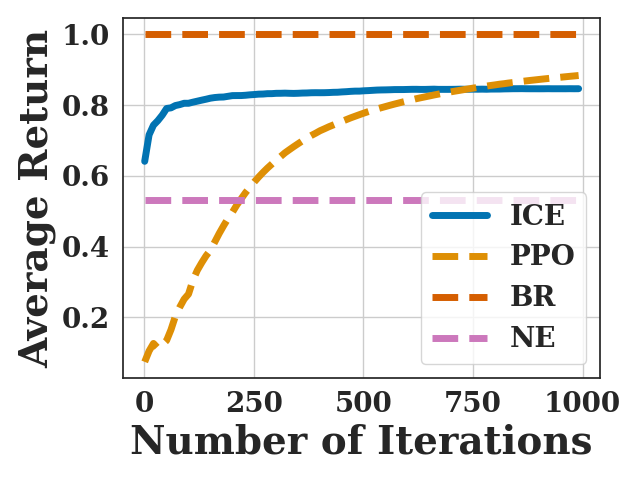} \includegraphics[width=0.32\columnwidth]{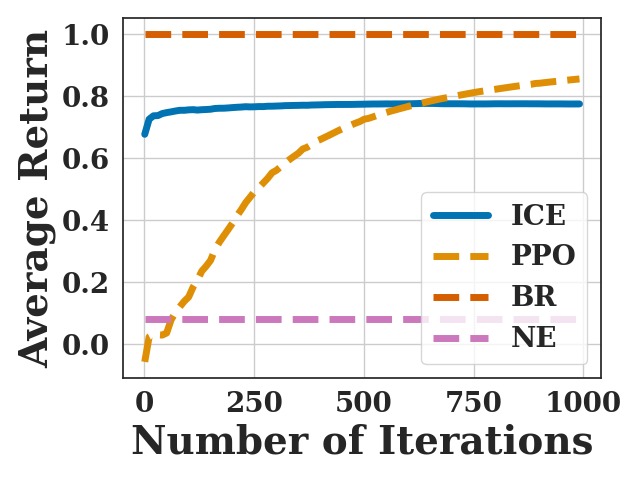}
     \includegraphics[width=0.32\columnwidth]{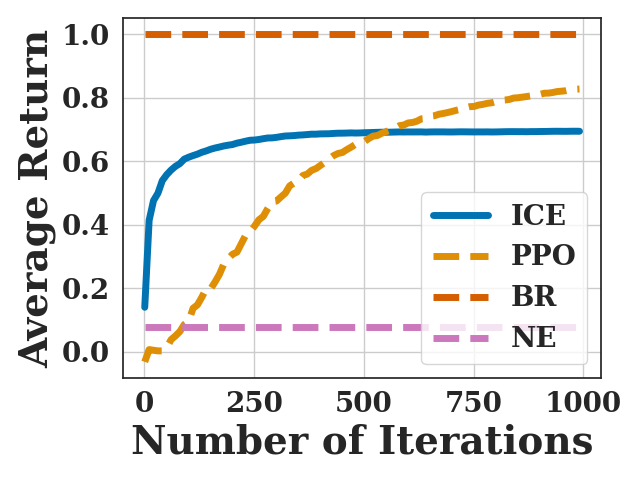}}
\caption{Out-of-distribution results of three-player games}
\label{fig:3_player_out_distribution}
\end{figure}

Next, Fig.\ref{fig:3_player_out_distribution} shows the results from the out-of-distribution testbed, where we observe trends similar to those in the in-distribution testbed.
The key distinction here is that the opponents in the out-of-distribution testbed are randomly generated, which often results in simpler strategic scenarios. In contrast, the in-distribution testbed encompasses a mix of randomly generated and learning-generated opponents, leading to potentially more complex and challenging interactions.
An interesting observation in the three-player Goofspiel game is that, after 500 interactions, the PPO algorithm begins to match the performance of ICE. This trend could be attributed to the simpler nature of the randomly generated opponents in the out-of-distribution testbed, which might be easier for PPO to adapt to and exploit over time. Despite this, ICE demonstrates a faster convergence to high-performance levels compared to the PPO algorithm and consistently outperforms the NE strategy. It indicates that ICE is not only capable of quickly adapting to new opponents but also effectively maximizing performance in diverse opponent settings, including both simple and complex strategic environments.

\begin{figure}[ht]
    \centering
    \subfigure[3-player Kuhn poker (left--player 1, middle--player 2, right--player 3)]{
    \includegraphics[width=0.32\columnwidth]{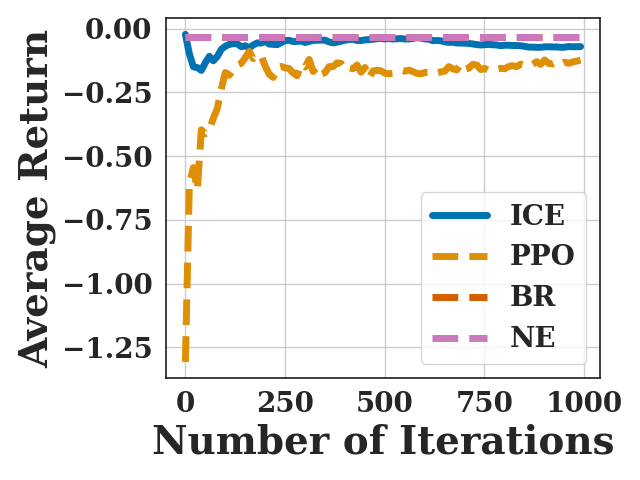}
\includegraphics[width=0.32\columnwidth]{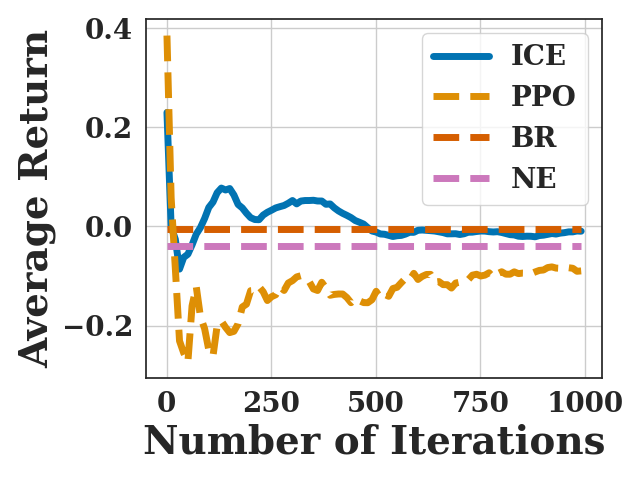}
\includegraphics[width=0.32\columnwidth]{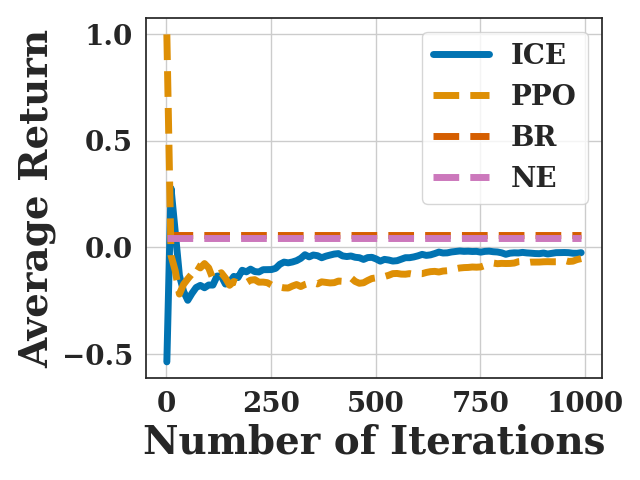}
     }
 \subfigure[3-player Leduc poker (left--player 1, middle--player 2, right--player 3)]{
     \includegraphics[width=0.32\columnwidth]{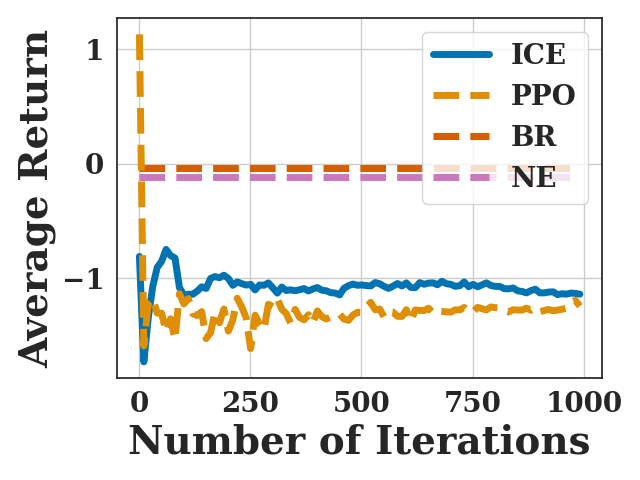} \includegraphics[width=0.32\columnwidth]{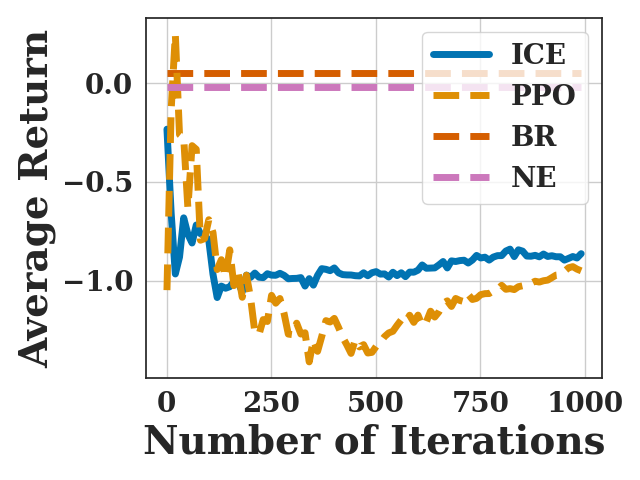}
     \includegraphics[width=0.32\columnwidth]{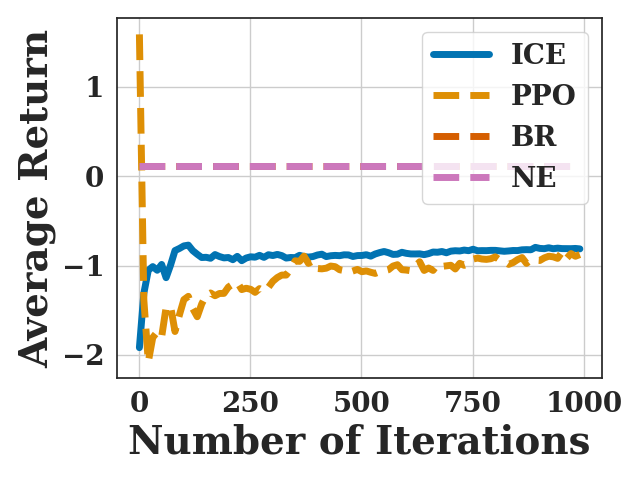}}
 \subfigure[3-player Goofspiel (left--player 1, middle--player 2, right--player 3)]{
     \includegraphics[width=0.32\columnwidth]{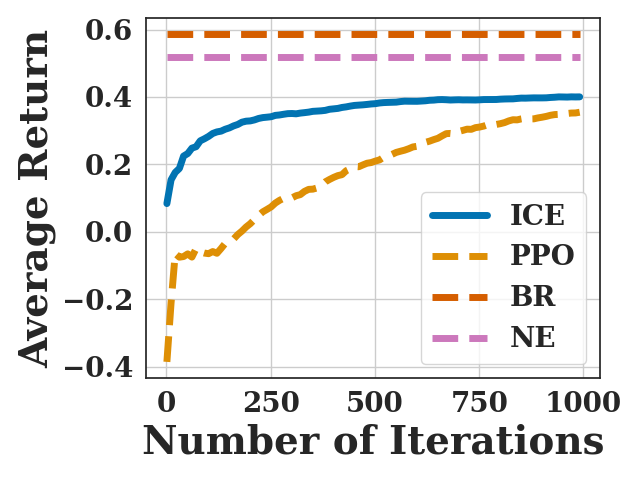} \includegraphics[width=0.32\columnwidth]{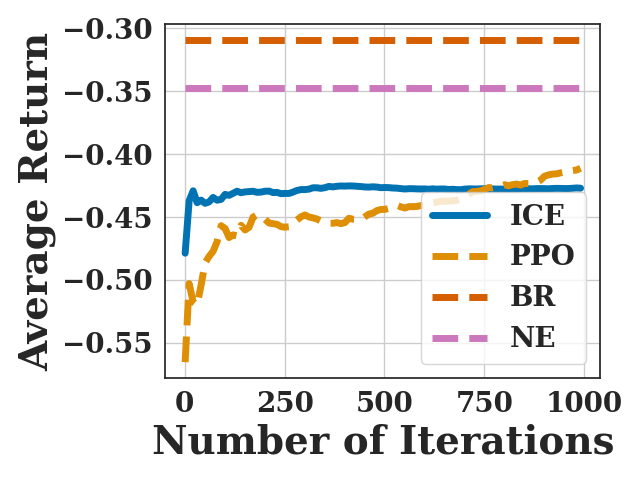}
     \includegraphics[width=0.32\columnwidth]{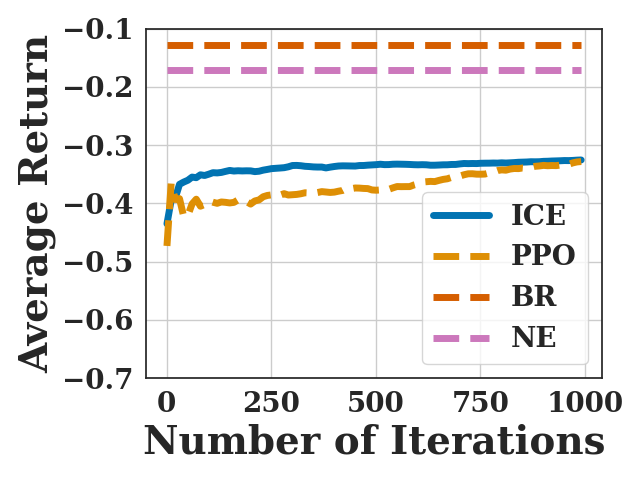}}
\caption{Results of three-player games against NE opponent}
\label{fig:3_player_ne}
\end{figure}

Lastly, we discuss the results against NE opponents, as shown in Fig.\ref{fig:3_player_ne}. Our findings reveal that the ICE algorithm achieves better or comparable performance to the NE strategy only in the three-player Kuhn poker game. However, in other game scenarios, while ICE does not outperform the NE strategy, it still maintains a higher level of performance than the PPO algorithm. The less optimal performance of ICE in these cases can be attributed to the highly dynamic game environment and stability of the NE opponents. In three-player games, the player faces two opponents simultaneously, and if both adopt the conservative NE strategy, exploiting them concurrently becomes significantly challenging. 
This observation highlights an area for future development. Improving the ICE algorithm to more effectively handle situations where multiple opponents employ highly conservative strategies, such as the NE, will be a focus of our future research.


\end{document}